# An Efficient Deep Learning-based approach for Recognizing Agricultural Pests in the Wild

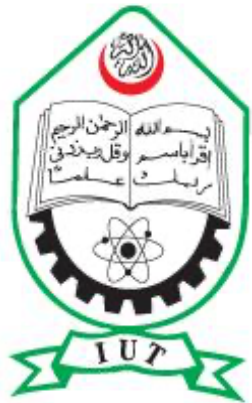

Authors:
Mohtasim Hadi Rafi
Mohammad Ratul Mahjabin
Md Sabbir Rahman

A thesis submitted for the degree of
**BSc in Software Engineering**

May, 2023

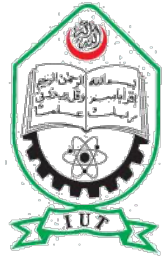

# Islamic University of Technology
Department of Computer Science and Engineering

# An Efficient Deep Learning-based approach for Recognizing Agricultural Pests in the Wild

Authors:

Mohammad Ratul Mahjabin, 180042123
Md Sabbir Rahman, 180042142
Mohtasim Hadi Rafi, 180042145

Supervisor:
Dr. Md. Hasanul Kabir
Professor, Department of Computer Science & Engineering,
Islamic University of Technology

Co-Supervisor:
Sabbir Ahmed
Assistant Professor, Department of Computer Science & Engineering,
Islamic University of Technology

May, 2023

# Certificate of Approval

The thesis titled, "An Efficient Deep Learning-based approach for Recognizing Agricultural Pests in the Wild" submitted by Mohammad Ratul Mahjabin, St No. 180042123, MD. Sabbir Rahman, St. No. 180042142, Mohtasim Hadi Rafi, St. No. 180042145 of Academic Year 2021-2022 has been found as satisfactory and accepted as partial fulfillment of the requirement for the degree Bachelor of Science in Software Engineering on April 20, 2023.

Board of Examiners:

Dr. Md. Hasanul Kabir
Professor
Department of Computer Science and Engineering
Islamic University of Technology (IUT), Gazipur

Sabbir Ahmed
Assistant Professor
Department of Computer Science and Engineering
Islamic University of Technology (IUT), Gazipur

# Declaration

It is hereby declared that-

- The report that is submitted is our original work at the time of completing our BSc in Software Engineering at Islamic University of Technology.

- The report does not include any already published or written by a third party content, unless it is properly cited in the references.

- The content of the thesis has not been previously approved or submitted for consideration for any other degree or certificate as our knowledge

- All significant sources of assistance have been acknowledged.

Student's Full Name & Signature:

Mohammad Ratul Mahjabin        Md Sabbir Rahman        Mohtasim Hadi Rafi
180042123                      180042142               180042145

Date: 20 May, 2023

ii

# Dedication

To the people who are working tirelessly to remove hunger from the world, who strive to provide sustenance to those in need, who believe that no one should go to bed hungry, this thesis is dedicated to you.

Your unwavering dedication to ending world hunger inspires us to do our part in creating a better future for all.

Thank you for your tireless efforts and commitment to this vital cause.

# Contents

# List of Figures

# List of Tables

# Acknowledgement

First of all, all praise to Allah swt for whom our thesis report have been com-pleted without any major interruption. Second, we would like to thank our supervisor, Dr. Md. Hasanul Kabir sir and co supervisor Sabbir Ahmed sir for their kind assistance and suggestions. They came to our aid anytime we needed it.

We also like to thank the staff, instructors, and students of Islamic University and Computer Vision Lab in the Department of Computer Science and Engineering.

It would not be able to carry out this work without the appropriate assistance of all these people. And finally, thanks to our parents, without whom we will not be able to be here.

We are currently preparing to graduate thanks to their kind prayers and sup-port.

[It is He] who has made for you the earth as a bed [spread out] and inserted therein for you roadways and sent down from the sky, rain and produced thereby categories of various plants.

- SURAH TA-HA AYAT 53 (20:53 QURAN)

Abstract

One of the biggest challenges that the farmers go through is to fight insect pests during agricultural product yields. The problem can be solved easily and avoid economic losses by taking timely preventive mea-sures. This requires identifying insect pests in an easy and effective man-ner. Most of the insect species have similarities between them. Without proper help from the agriculturist academician it's very challenging for the farmers to identify the crop pests accurately. To address this issue we have done extensive experiments considering different methods to find out the best method among all. This paper presents a detailed overview of the experiments done on mainly a robust dataset named IP102 in-cluding transfer learning + finetuning, attention mechanism and custom architecture. Some example from another dataset D0 is also shown to show robustness of our experimented techniques.

Keywords: Deep Learning, Transfer learning, Deep Neural Network, Pretrained Model, IP102

# 1 Introduction

## 1.1 Motivation

The escalating global population and the ever-increasing demand for food have underscored the urgent need to enhance agricultural productivity while effectively managing the impact of agricultural pests on crop yields and quality. One significant factor that hampers agricultural productivity is the damage caused by insect pests to crops worldwide each year. Preventing such damages is crucial for commercial benefits, improved agricultural efficiency, and avoiding significant harvest losses. In addition to reducing the yield, crop pests can also inflict damage on machinery, equipment, soil, and infrastructure [18]. Conventional methods employed to identify and classify agricultural pests in the wild are often characterized by time-consuming and expensive processes that heavily rely on expert knowledge. As a result, there is a compelling necessity for the development of an innovative and efficient approach to address this challenge.

By mitigating these problems, agriculture can drive economic growth while producing food with reduced resource consumption. Consequently, the recogni-tion and classification of pests assume paramount importance in preventing crop damage. In recent years, there has been a notable surge of interest in automatic pests' classification [44]. Studies have shown that early detection and treatment can minimize damage to almost zero. However, this task is far from easy, as cur-rent methods for identifying insect pests are inefficient and expensive, relying heavily on the technical expertise of agricultural professionals.

Deep learning technology has emerged as a powerful tool in various fields, showcasing remarkable achievements, particularly in image-based tasks. This technology offers a promising opportunity to create a highly accurate and scal-able solution for the recognition and classification of agricultural pests in diverse environmental conditions. By leveraging deep learning algorithms, it becomes possible to train models that can effectively analyze and interpret complex vi-sual data, enabling accurate identification and classification of pests present in agricultural settings.

Exploiting deep learning technology offers immense potential in revolution-izing pest classification by providing efficient and cost-effective solutions. By leveraging the power of deep learning algorithms, researchers can develop ro-bust models capable of accurately identifying and classifying insect pests, even in large-scale agricultural settings. The utilization of deep learning techniques can enhance the speed, accuracy, and scalability of pest detection and enable early intervention to minimize crop damage effectively.

Fortunately, advancements in deep learning techniques have paved the way for their adaptation across various domains, yielding remarkable results. Deep learning has achieved significant breakthroughs, particularly in image-based computer vision tasks such as image classification, segmentation, and detection [34, 14, 15]. The field of agriculture has also witnessed the successful applica-tion of deep learning methods in diverse areas, including plant identification, recognition, and classification [34, 9, 48, 16, 24], fruit counting [6], and plant disease detection [28].

As the world grapples with the challenge of feeding an ever-growing population, ensuring food safety becomes imperative. Addressing the issue of insect pests, a major cause of crop damage, is crucial for improving agricultural productivity. Deep learning technology, with its significant achievements in image-based computer vision tasks, holds tremendous promise in the field of agriculture. By harnessing deep learning algorithms, researchers can enhance the recognition and classification of pests, leading to more efficient and cost-effective solutions for safeguarding global food production.

## 1.2 Problem Statement

Our research endeavors encompassed a comprehensive array of experiments con-ducted on the esteemed IP102 dataset [43], which is widely acknowledged in the field of agricultural pest recognition and classification. This dataset posed multifaceted challenges that required meticulous attention to detail in order to derive robust solutions. Notably, these challenges included the presence of pests within intricate backgrounds and foregrounds, the remarkable variability in their size, shape, color, and texture, and the marked imbalance in sample distribution among different pest species.

To surmount these hurdles and fortify the efficacy of our models, we devised a series of astute strategies that leveraged cutting-edge techniques. In partic-ular, we delved into the realm of combined feature extraction, deftly merging ConvNext, an acclaimed architecture, with the innovative Vision Transformer (ViT) feature extractor. This synergistic amalgamation enabled us to har-ness the distinctive strengths of both architectures, culminating in an enhanced ability to capture a diverse range of features that are crucial for discerning and distinguishing between different pests with utmost precision.

Furthermore, we meticulously fine-tuned the linear dropout hyperparameter, painstakingly seeking an optimal equilibrium that would effectively mitigate the risks of overfitting while concurrently amplifying the generalization prowess of our models. This nuanced fine-tuning process imbued our models with a refined aptitude for accommodating diverse and unseen data samples with remarkable finesse.

Addressing the challenges emanating from complex backgrounds and fore-grounds necessitated a meticulous exploration of transfer learning and fine-tuning methodologies. By adroitly harnessing the immense knowledge encapsulated within pre-trained models and adorning them with specialized adaptations to suit the intricacies of agricultural pest classification, we attained unprecedented strides in performance optimization. Furthermore, we judiciously employed ingenious data augmentation techniques such as Fmix and Cutmix to ingeniously alleviate the prevailing data/class imbalance and to augment the richness and diversity of our training samples. Additionally, we harnessed the potential of a cropped dataset, expertly extracting the regions of interest (ROI), and skillfully employing segmentation techniques to meticulously isolate and focus on the pivotal pest-related regions within the images.

The significance of attention-based methods was profoundly underscored in our research endeavors. We fervently explored multiple CNN-based models, adroitly integrating attention mechanisms into the core fabric of our models. A quintessential embodiment of this was manifested through the incorporation of the Convolution Block Attention Module (CBAM), an avant-garde innovation that facilitated an adaptive recalibration of feature maps, enabling our models to dynamically emphasize and prioritize the salient regions relevant to the pests under consideration. This judicious emphasis on pertinent information translated into a remarkable enhancement in the accuracy and acuity of our pest classification endeavors.

In our relentless quest for superior performance and minimized generalization errors, we judiciously employed ensemble methods, an ingenious ap-proach that synergistically harnessed the collective wisdom of multiple models. Through astute employment of soft voting and hard voting mechanisms, we deftly amalgamated the predictions generated by the diverse array of models at our disposal. This ensemble-based confluence of perspectives endowed our mod-els with heightened confidence and decisiveness, thereby effectively curtailing the risks associated with generalization errors and fostering superior classifica-tion outcomes.

## 1.3 Organization

Moreover, our research encompassed an incisive examination of the challenges posed by intra-class dissimilarity and inter-class similarity. By deftly strati-fying pests based on their distinct life stages, namely early and late stages, we embarked upon a revelatory exploration of the metamorphic nature exhib-ited by most pests. This compelling insight enabled us to not only discern the prominent intra-class dissimilarities but also delineate the striking inter-class similarities that surfaced when considering pests throughout their life cycles. This meticulous classification based on pest stages served as a powerful tool for enhancing classification accuracy and, simultaneously, provided invaluable insights into the developmental. This insightful classification based on pest stages served as a powerful tool for enhancing classification accuracy and pro-viding valuable insights into the developmental aspects of pests [16, 24].

We have divided the paper into five more sections. The second section con-sists of literature review. We have read and reviewed other research works done in recent years in this domain and tried to find out the research problems. In the third section, namely methodology, we have discussed about our experi-ments and proposed methods. The fourth section consists of the results of our experiments and we tried comparing our results with other literature to find out where we stand. Finally we draw a conclusion to the paper and discussed what are the impacts of this work.

## 2   Literature review

Many researchers have used deep learning algorithms in agriculture in recent years. GoogleNet, AlexNet, ResNet, and other deep convolutional neural net-works perform admirably well for image classification tasks. The majority of the applications, however, are focused on weed identification [9] plant recognition [34, 48], fruit counting [6], and crop type [48] classification.

### 2.1   Subdomains in agriculture

Research in agriculture is important because it helps to improve the efficiency, productivity, and sustainability of agricultural practices. By studying the best ways to grow crops, raise livestock and fisheries, and manage pests and diseases, researchers can help to increase crop yields, reduce the need for chemical inputs, and protect the environment. The agriculture field is very vast there are a number of domains to work on here. Kamilaris and Prenafeta-Boldu [17] and Saleem et al [35] have identified the subdomains of the agriculture domain as follows-

- Plant leaf disease detection, segmentation and diagnosis
- Plant disease detection and diagnosis
- Crop pest detection
- Crop analysis with aerial image
- Crop weed detection
- Crop yield (harvest) prediction
- Crop classification
- Crop growth monitoring

### 2.2   Use of deep learning and it's influencing factors

Different machine learning techniques name as Regression, Clustering, Bayessian Models, Instance Based Models, Decision Trees, SVM, KNN etc are being used already in this domain [23]. These techniques had put some impact results in the research of agriculture system automation but the real evolution came when deep learning architecture took the place. Now networks can be more deeper and a lot of resources can be used to feed the networks. That shows tremen-dous advancement in the field of computer vision based agricultural research. Researchers now using different deep learning networks like CNN, GoogleNet, Resnet, VGG , Inception etc [17]. As agriculutre based sectors are based on mostly image and video clips. Deep learning techniques has a large impact on this. After studying some survey papers we narrowed down the field by select-ing two sub sectors which seemed to have more potential. The fields are plants' disease detection, segmentation & diagnosis and crop pest classification. First we conducted studies on plant disease detection. There are some factors which influences highly for deep learning based plant disease recognition[4]-

- Limited annoted dataset: There are not enough qualityfull dataset in this agriculture field. Making a good dataset in this field not so easy. First of all, enough sample should be ensured to feed the deep learning networks. Secondly data annotation should be done carefully because it is very difficult for an ordinary person to identify different disease or crops of agriculture field. Moreover these disease or crops may be at the different growth stage of the life.

- System representation: There are a lot of types and varieties are available in this field. There are different species and classes for a single plant. Moreover a same plant may look different based on different conditions and environments. Without this there may be a similar disease or pest which can affect multiple plant species. So a good representation of these info in the dataset is very important which is a rare case.

- Covariate shift: At the time of training and evaluating a method on the same dataset, the model's performance is often overstated because it will fail when implemented to other datasets. As previously stated, datasets are not robust, and even the same plant can appear differently in different locations, so models frequently fail to perform well for images other than their own testing and training datasets.

- Image background: Complex background is a big problem in this sec-tor. For example a single disease may effect different plant but the color, size in the background can be different. Moreover there will be soil , grass, different plants hands etc present in the picture. Often the focused object like disease zone or pest can be differ in size. There are also huge possibility of color similarity between the focused objects and the background. The focused leaf or crop can be damged differently also for different which can cause change in the background.

- Image capture condition: Images captured in most of the datasets differ from one to another depending on the condition of the day, light time, camera positioning, environment and so many. Due to the reason, region of interest looks different from one image to another in the same class or even same insect of the same dataset.

- Symptom segmentation: The fact that it is not essential to precisely identify the indicators in the image is one of the key benefits of employing deep learning methods.The issue of symptom segmentation is not pertinent in this situation, despite the fact that it may be useful to isolate the area where the symptom is present because it often contains the majority of the crucial information. The results were not significantly changed by expanding the sample size or by limiting the study area to the site of the symptoms. Furthermore, by concentrating on specific nodules, it is feasible to integrate the predicted classes to develop a comprehensive diagnostic for the plant, which might lessen the impact of individual mis-classifications.

- Symptom variations: While the majority of diseases have recognizable visual symptoms, these symptoms frequently vary in appearance, especially in terms of color, shape, and size. Due to this unpredictability, it may be difficult to employ visible-spectrum-based image-based diagnos-tics to discriminate between healthy and unhealthy pixels. [4]. As symp-toms can range from being extremely faint and barely perceptible in the early stages of infection to causing extensive tissue death in the most advanced stages, the stage of the disease (or the intensity of the symptoms) is probably the most significant source of heterogeneity. This means that differentiating various diseases may be simpler or more difficult depending on the stage of infection.

- Simultaneous disorders: Images typically have an illness tagged next to them. However, it is typical for several illnesses or other types of issues, like nutrient deficits or pests, to exist concurrently. This is frequently true because weakened by one virus, a plant's immune system can make it more vulnerable to other diseases. Creating mixed classes with all conceivable combinations of disorders could be one solution to this problem. This strategy is not realistic, though, as there would be an excessive number of classifications, which would raise the possibility of misclassification. Additionally, the inter-variability would be too high because the percentage of symptoms linked to each disease can differ from one image to the next.

- Disorders with similar symptom: One host species may have lesions and other symptoms from a variety of agents, such as illnesses, nutri-tional deficits, pests, and mechanical harm. Even plant pathologists have faced trouble sometimes for differentiating between some of these chem-icals since their symptoms can be so similar. As a result, visual cues alone might not be adequate to correctly categorize some issues. Even if an image is very clearly caught, a specific diagnosis could be impossible because of the symptoms' broad nature. Human experts frequently take into account additional facts to make accurate decisions, such as the current weather condition, historical illness data and stats, and the overall health of the plant. The accuracy of disease recognition algorithms may be increased by including this kind of additional data. [17].

## 2.3 Existing works

Mohanty et al. [28] trained AlexNet and GoogleNet using PlantVillage dataset which contains 54,306 images with 14 crop species and 26 kinds of diseases (only lab environment image) via transfer learning. A novel CNN-Fourier Dense

Network was proposed and evaluated by Lin et al. [24] with their self-built dataset based on the optical images captured using an unmanned aerial vehicle.

Wang et al. [41] applied AlexNet and LeNet deep networks and achieved a classification accuracy of 91%. They used their self-made dataset containing 30,000 pest images in 82 classes and also analyzed the kernels effect in the cnn layers and cnn layers number on the classification performance.

Wu et al. [43] collected image data from different sources like internet, newspaper, magazine etc and created a large dataset of insect pests which contains more than 75,000 field images belonging 102 categories of crops where about 19,000 box annotated images for object detection. The dataset is evaluated on some classical machine learning techniques and also modern deep learning-based techniques. Later many used the database to evaluate their own approaches. Such includes Ren et al. [33]. They came up with a new feature reuse residual block (FR-ResNet) structure which is based on classic residual blocks which can help to improve the capacity of data representation. With their proposed technique they achieved an accuracy of 55.24% using FR-ResNet whereas the state of the art ResNet-50 method brings 54.19% accuracy. Liu et al. [25] also proposed a new residual-based block network architetcure named multi-branch fusion residual network (DMF-ResNet) for multi-scale representations. In the proposed method conventional residual network is combined with bottleneck residual network architecture into the residual model with multiple branches. They measured the performance with other sota methods and the experiment came up with an enhancement in the result.

Nanni et al. [29] also used IP102 and another small dataset for their proposed ensemble strategy which combines saliency methods and CNNs. They augmented the data using different saliency methods and able to achieve an accuracy of 91% in their small dataset. But the accuracy in IP102 was 61.93%. Ayan et al. [3] proposed a genetic algorithm based weighted ensemble of deep convolutional neural networks. They used D0, a small dataset with 40 classes and IP102 to evaluate their performance and achieved 98.19% for the D0 dataset, 95.15% for a dataset created by taking 10 classes of IP102 and 67.13% in IP102. Both of the approaches are enhanced compared to the sota methods of classifications but full ip102 dataset is not taken.

# 3 Methodology

## 3.1 Dataset

Insect pests evolve and change their visuals during all their lifetime depending on the species and category. Collecting and classifying pest images becomes more difficult due to these reasons.

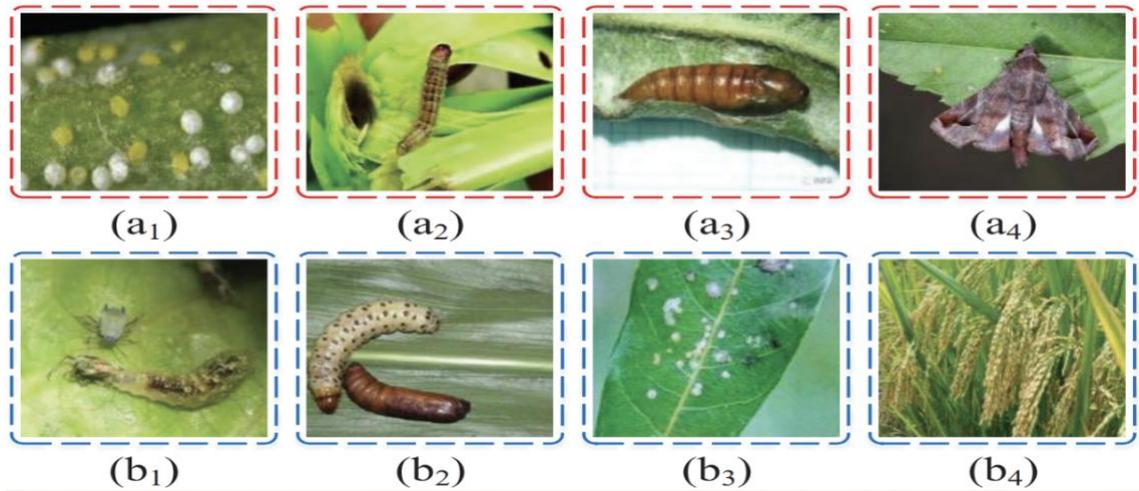

Figure 1: IP102 Sample Images of different growthstage and multiple samples [43]

Hence we have used a public dataset named IP102. It is a large dataset con-taining more than 75,000 images belonging to 102 classes. Some sample images from the dataset are shown in fig 1. This dataset is mainly devided into two categories- field crops and Economic crops. Field crops consist of five super classes- Rice, Corn, Wheat, Beet, Alfalfa. Economic crops consist of three super classes- Vitis, Citrus, Mango.

Hence we have used a public dataset named IP102. It is a large dataset containing more than 75,000 images belonging to 102 classes. Some sample images from the dataset are shown in fig 1. This dataset is mainly devided into two categories- field crops and Economic crops. Field crops consist of five super classes- Rice, Corn, Wheat, Beet, Alfalfa. Economic crops consist of three super classes- Vitis, Citrus, Mango.

Hence we have used a public dataset named IP102. It is a large dataset containing more than 75,000 images belonging to 102 classes. Some sample images from the dataset are shown in fig 1. This dataset is mainly devided into two categories- field crops and Economic crops. Field crops consist of five super classes- Rice, Corn, Wheat, Beet, Alfalfa. Economic crops consist of three super classes- Vitis, Citrus, Mango.

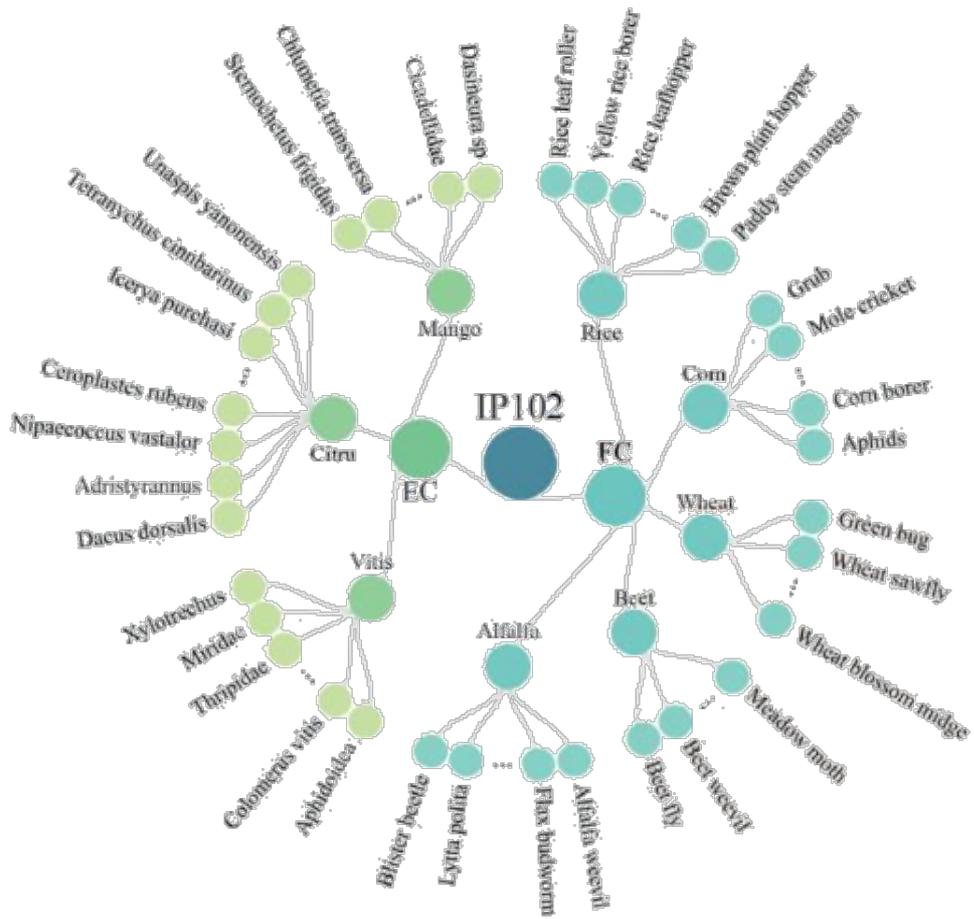

Figure 2: Taxonomy of the IP102 dataset [43]

| Superclass | Class | Train | Val | Test | Total sample / superclass | IR* |
|---|---|---|---|---|---|---|
| RC* | Rice | 14 | 5,043 | 843 | 2,531 | 8,417 | 6.4 |
| | Corn | 13 | 8,404 | 1,399 | 4,212 | 14,015 | 27.9 |
| | Wheat | 9 | 2,048 | 340 | 1,030 | 3,418 | 5.3 |
| | Beet | 8 | 2,649 | 441 | 1,330 | 4,420 | 15.4 |
| | Alfalfa | 13 | 6,230 | 1,037 | 3,123 | 10,390 | 10.7 |
| EC* | Vitis | 16 | 10,525 | 1,752 | 5,274 | 17,55 | 74.8 |
| | Citrus | 19 | 4,356 | 725 | 2,192 | 7,273 | 17.6 |
| | Mango | 10 | 5,840 | 971 | 2,927 | 9,738 | 61.7 |
| IP102 | FC | 57 | 24,374 | 4,060 | 12,226 | 40,660 | 39.4 |
| | EC | 45 | 20,721 | 3,448 | 10,393 | 34,562 | 80.8 |
| Total | IP102 | 102 | 45,095 | 7,508 | 22,619 | 75,222 | 80.8 |

Table 1: IP102 Dataset
FC = Field crops, EC = Economicalcrops, IR = Imbalance Ratio

## 3.2 Data preparation, Pre-processing and Training procedure

Millions of parameters are worked on a neural network system. For better gener-alization capability it is essential to perform data augmentation in the dataset. All the images of the dataset are resized to 224 * 224 * 3 and RGB format for performance betterment of the system. The data are kept with maintaining proportion so that a good learning capacity can be achieved. Different aug-mentation like horizontal flipping, vertical flipping, 10°rotation and slant angle (0.2) for sheer transformation is done to both training and validation data to make up for the lack of data availability. Different types of inputs are being used to validate the model. For that, the validation data portion is also aug-mented. Both synthetically modified data and validated augmented data are required for the learning of the models. The images are normalized by divid-ing the images using RGB channel mean and calculating standard deviation of the images in the ImageNet1K dataset ([0.485, 0.456, 0.406] and [0.229, 0.224, 0.225]) so that uniform data distribution can be ensured. This also ensures bet-ter convergence of the model during the training of the neural network. Dataset splitting for train, validation, and testing has been maintained with a 6:1:3 ratio. For training all images are normalized to the . For all models, the final linear classification layer is replaced with a new layer with as many output nodes as there are classes to classify the dataset, and model parameters are optimized so that the categorical cross-entropy loss function is minimized. If we consider the probabilities of the events from P and Q, then cross-entropy can be calculated as-

$$H(P, Q) = -\sum_{x \in X} P(x) * \log(Q(x)) \qquad (1)$$

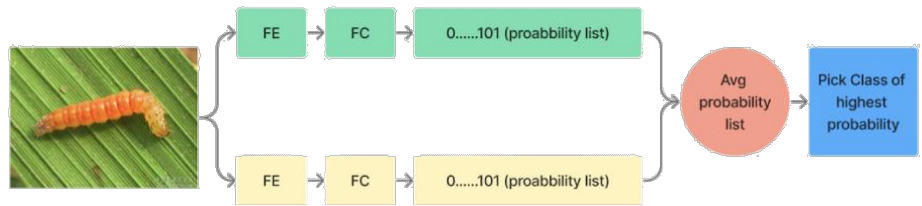

Figure 3: Ensemble Technique

## 3.3 Experimental Setup

Colab and Kaggle Notebooks were used to generate a python environment with Pytorch and other python libraries to carry out the experiment of the proposed method. An Intel Xeon CPU @ 2.00 GHz, thirteen(13) GB Ram and a Tesla P100 16GB VRAM as GPU were used to conduct most of the experiments. On the other hand we also used high config Intel Core I9 pc with 24 gb nvdia rtx 3080 for some heavy computation task like segmentation.

For all the classes, the sample images were split into a 6:3:1 ration for train-ing, validation and testing. Batch size 32 was selected for mini batch gradient descent. Early stopping was used which help to reduce overfitting problem and improve generalization of the model. Each pretrained model was set to train for at least 25 epoch with early stopping. In our method, a change of $10^{-4}$ is considered great. Otherwise it is considered a not improving epoch. The setting was made to stop training early if the training completes ten consecutive non imroving epochs. On average most of the models were able to converge after 20 epochs but two of them got early convergence after 10 epochs. For optimization Adam optimizer was used as it is usually recommended for classification tasks. Adam optimizer has faster computation time and fewer parameters tuning.

Since it is a multiclass classification task, cross entropy loss was used to calculate the loss.Firstly, the learning rate was set to 0.001 and after every seven non improving epochs, the learning rate was set to decrease by a factor of 0.1 to assist the model for finding a set of globally optimal weights that improve generalization. In the experiment, pretrained models were initialized using ImageNet dataset's weight. Model checkpoints were used to save the model with the best validation accuracy so that the training can be continued later from the point it was stopped.

## 3.4 Experiment Methods

### 3.4.1 Convolutional Neural Networks

Convolutional Neural Networks [21] (CNNs) have emerged as a prominent and extensively employed deep learning methodology within the domains of computer vision and image processing. CNNs have brought about a paradigm shift in a transformative era in handling image-related tasks, such as image classi-fication, object detection, and image segmentation. A fundamental strength inherent in CNNs is their innate capacity to automatically acquire and extract hierarchical features from unprocessed input data. By effectively utilizing convolutional layers, pooling operations, and non-linear activation functions, CNNs adeptly capture spatial representation and meticulously preserve vital structural information embedded within images. The multi-layer architecture of CNNs enables them to learn complex patterns and representations, making them highly suitable for tasks involving large-scale image datasets. As a result, CNNs have achieved remarkable performance improvements in image recogni-tion tasks, surpassing traditional machine learning methods and setting new benchmarks

### 3.4.2 Transfer learning and Fine Tuning

Transfer learning is where there are pretrained models which are previously trained on a big dataset and then the knowledge is transferred into the new desired dataset using fine tuning. Transfer learning [1] leverages the idea that knowledge acquired from solving one task can be beneficial for solving a related but different task. Instead of training a model from scratch, transfer learning starts with a pre-trained model that has been trained on a large-scale dataset, typically on a source task or domain. The knowledge captured in the pre-trained model's parameters, also known as weights, can be transferred and utilized to enhance the learning process for the target task or domain utilizing fine tuning on the smaller dataset [10]. In fine tuning the last few layers are trained again for the new dataset but not the whole architecture. It saves time and computational effort as well. It's been very effective on the classification task. So we firstly tried different pretrained models such as ResNet, EfficientNet and so on. The last fully connected layer was changed to the class number for IP102 which is 102. Then the model was trained on the IP102 dataset.

Pretrained Models

- EfficientNetV2:EfficientNets [37] are a series of CNNs that have achieved excellent results on the ImageNet [20] dataset. They are designed using a method called compound scaling, which involves adjusting the width, depth, and resolution of the model. The optimal values for these factors are determined using a technique called Neural Architecture Search, which aims to minimize floating-point operations number (FLOPS) required. A newer version of EfficientNets called EfficientNetV2 [37] has been developed, which is smaller and faster for classification tasks. It uses a different type of block called Fused-MBConv, which includes a standard convolution with 3x3 filters rather than a depthwise convolution. To the best of the authors' knowledge, there has not been any previous work on trans-fer learning using the EfficientNetV2 model. The authors have conducted their own experiments with transfer learning on EfficientNetV2 and found that it performs very well for classifying images as normal or abnormal.

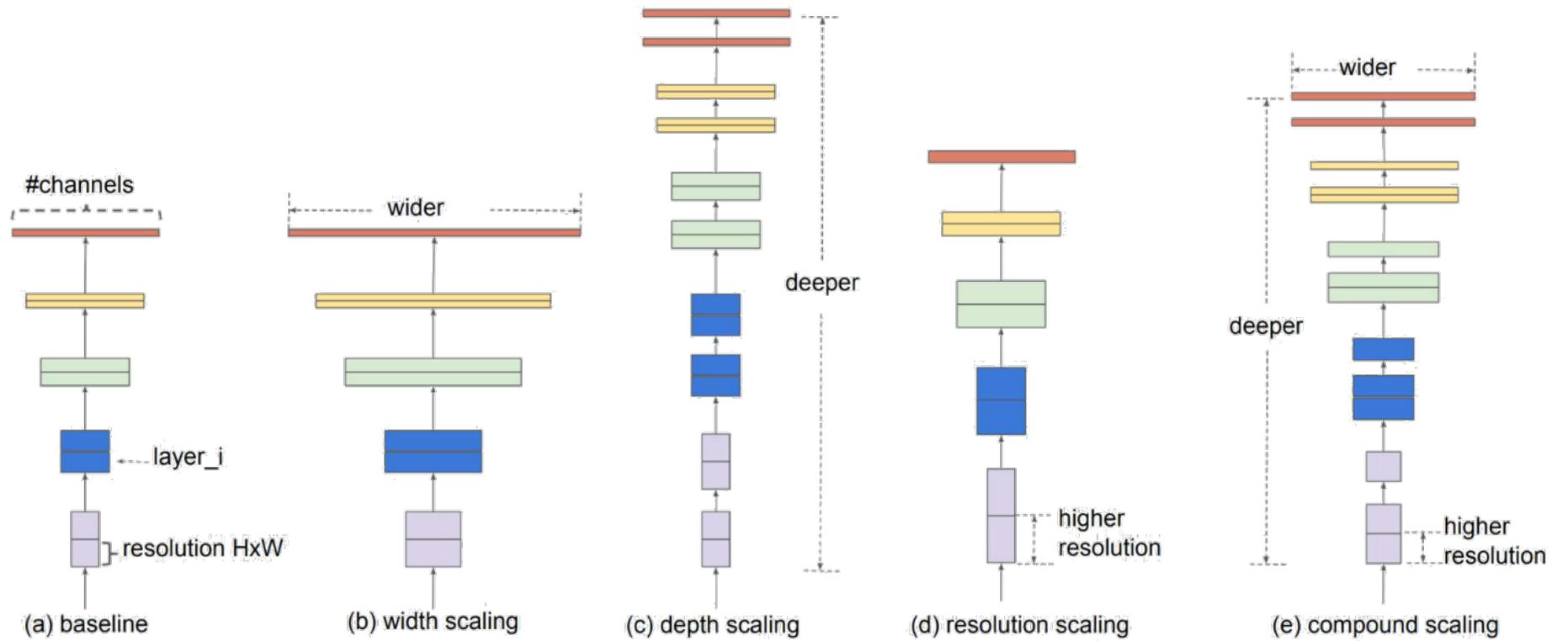

Figure 4: Model Scalling of Efficient Net [37]

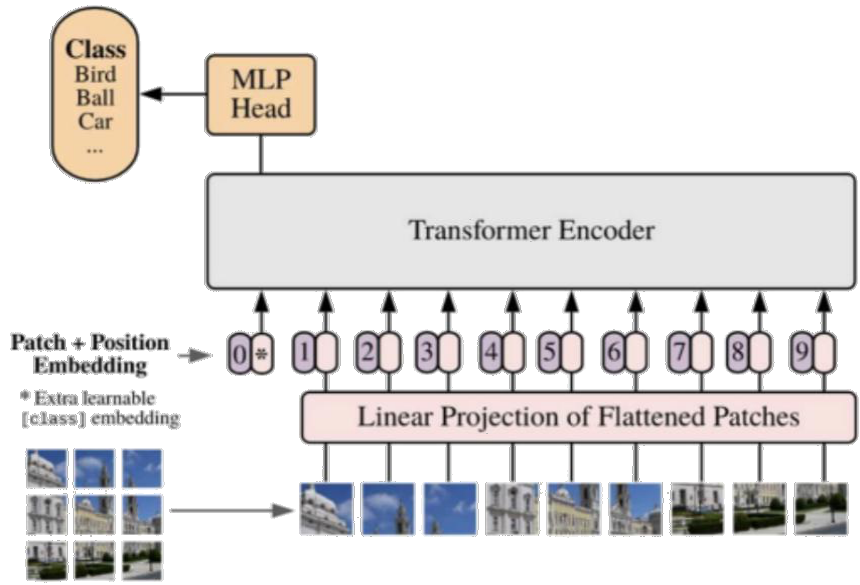

Figure 5: Encoder Model of ViT Network [31]

In this report, the ConvNeXt-L model was trained by using its default configuration . It was pretrained with the ImageNet21K dataset and has 235M parameters.

- ConvNeXt: ConvNeXt is a deep learning framework used for semantic segmentation and object detection. The attentional mechanisms are absent. It makes advantage of transformer networks by "modernizing" the ResNet network (ResNeXt) [13] ResNeXt is employed rather than ResNet, despite ConvNeXt's resemblance to the Swin Transformer [26] model. Transformer networks make use of advances in the ConvNeXt block (e.g. AdamW optimizer). Figure 6 illustrates the ConvNeXt block, which includes the convolution layer, Linear Normalization with Gaussian Error Linear Unit (GELU).

- Vision Transformer (ViT): In the field of deep learning, attention mechanisms are a recent development that are particularly useful for natural language processing tasks. The ViT model was the first to utilize this technique in image segmentation. It works by dividing the image into smaller pieces and encoding them with position values, which are then passed to the transformer decoder for classification. By using atten-tion mechanisms, the ViT model is able to better understand and analyze images for accurate segmentation and classification.

Figure 5 illustrates the encoder model of the vision transformer (ViT), which includes k self-attention mechanisms (also known as multihead self-attention). These mechanisms are calculated where the query matrix, key matrix, and value matrix are used to determine the attention given to each element. The multi-headed self-attention mechanism, shown in Equation 3, is made up of multiple self-attention operations and was pre-trained with ImageNet21K. In the ViT architecture, these mechanisms are used to analyze and understand the input image in order to perform accurate segmentation and classification.

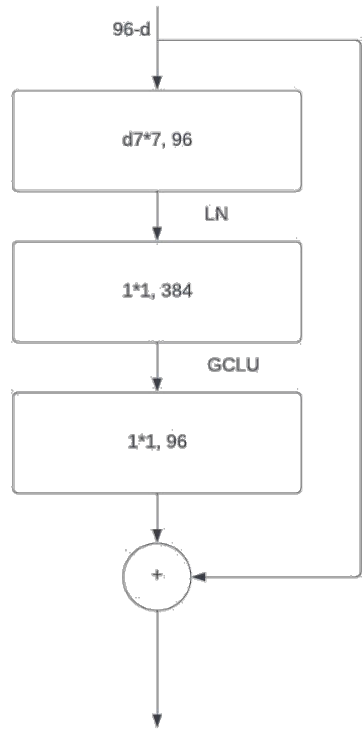

Figure 6: ConvNext Block [31]

### 3.4.3 Attention Mechanism

After examining the results of different pretrained models in the datasets of IP102, it is observed that the dataset is a difficult one as it has images with complex background, wild images and even sketches of pests as well. Hence the dataset gives the pretrained model a hard time to learn properly. As a result the accuracy obtained is in the range of 70 76% which is not that satisfactory for a classification task. so we figured out that the model was not really giving attention to the portion of the images where it should give as the insects take only a small portion of the whole image. so after we implemented the idea of the literature [39]. They presented different cnn models in their work with some attention mechanism like RAN, feature pyramid(FAN) and ensemble them to get a better result on IP102.

Residual Networks: Residual networks or ResNet [13] was proposed by He et al. in 2015. This is a network architecture that utilizes skip connections, or shortcut pathways, between layers. These skip connections allow the gra-dient to flow back to the input more easily, which helps prevent the vanishing gradient problem that can occur when training deep neural networks. This allows the network's weights to be updated more effectively. Residual networks are made up of residual blocks that can be stacked to create very deep neural networks, potentially with over 1000 layers, depending on the specific problem being addressed. These networks are depicted in figure 7.

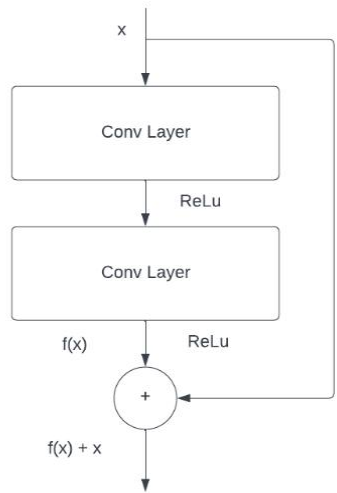

Figure 7: Structure of a residual block in residual networks [39]

Residual Attention Networks (RAN): Wang et al. [40] introduced a method called (RAN) that utilizes attention mechanisms within convolutional neural networks (CNNs) to identify important areas of an image for classi-fication. The Residual Attention Network is designed by layering numerous Attention Modules. Each of these modules comprises two parts: the mask branch and the trunk branch. The trunk branch handles feature processing and can integrate various network structures, such as pre-activation Residual Unit, ResNeXt, and Inception, used as the fundamental unit of the Residual Atten-tion Network.The mask branch, on the other hand, uses a bottom-up top-down structure to create a mask of the same size, M(x), that softly weights output features from the trunk branch, T(x). This mask functions as control gates for the trunk branch neurons, similar to the Highway Network. The attention mask not only acts as a feature selector but also guides gradient updates during backpropagation, enhancing the module's robustness to noisy labels. A unique aspect of the Residual Attention Network is that each trunk branch within an Attention Module has its own mask branch, which learns attention specifically for its features. This enables more efficient refinement and processing of com-plex images. These networks are composed of multiple attention based modules that can generate attention-based features to guide the learning process, which results improved performance compared to previous methods. Each attention module has two parts: one is trunk branch for feature extraction, which can be adapted to any network structure, and another is a mask branch that learns attention masks to weigh the output features and select relevant ones for clas-sification. The output of an attention module with residual-attention learning can be represented by the following equation -

$$H_{i,c}(x) = (1 + M_{i,c}(x)) * F_{i,c}(x) \qquad (2)$$

here x means input, i means ranges over all spatial positions, and c means the index of the channel ($c \in 1, ..., C$). Also, M(x) is denoting the mask branch output and on the other hand F(x) is the original extracted feature by the trunk branch.

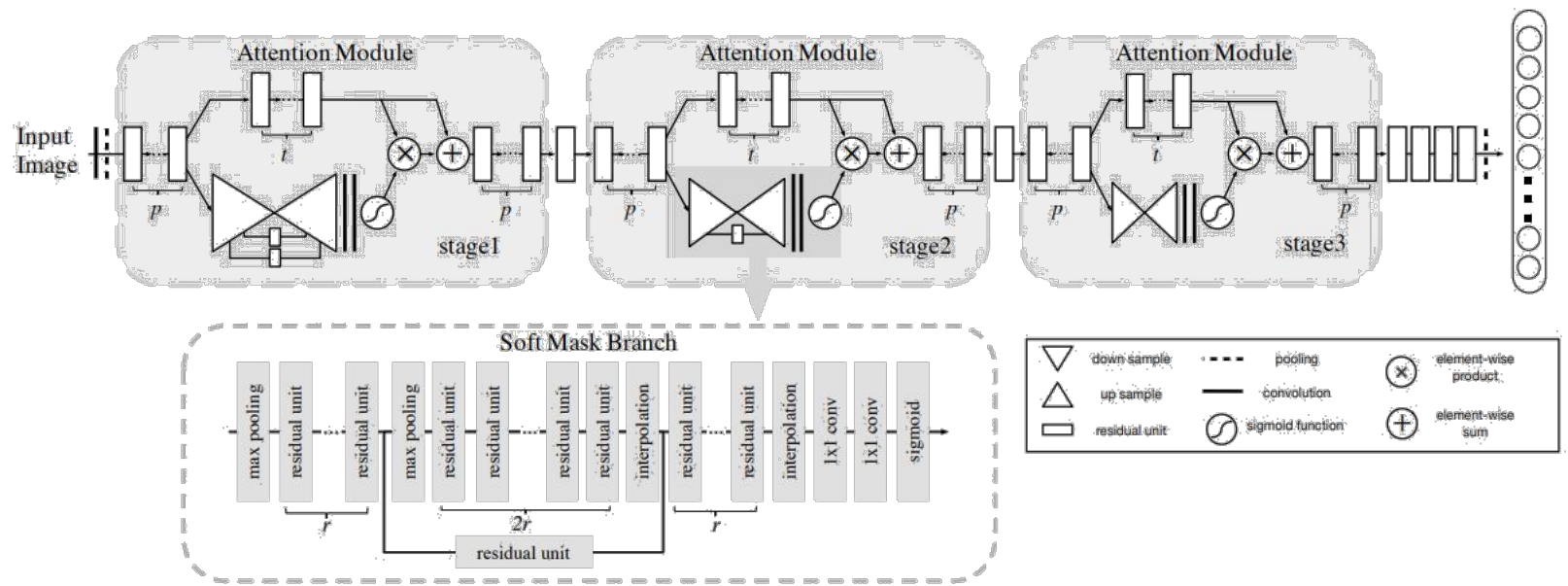

Figure 8: Residual Attention Network [40]

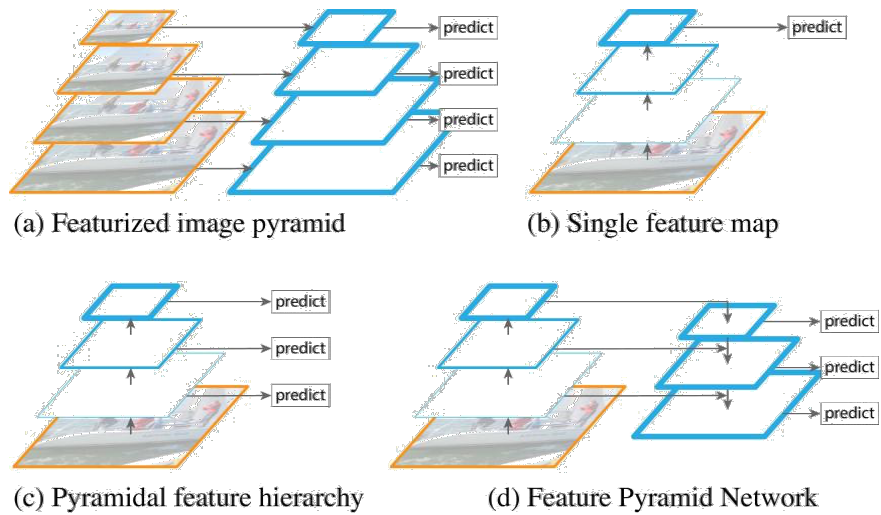

Figure 9: Feature pyramid networks [39]

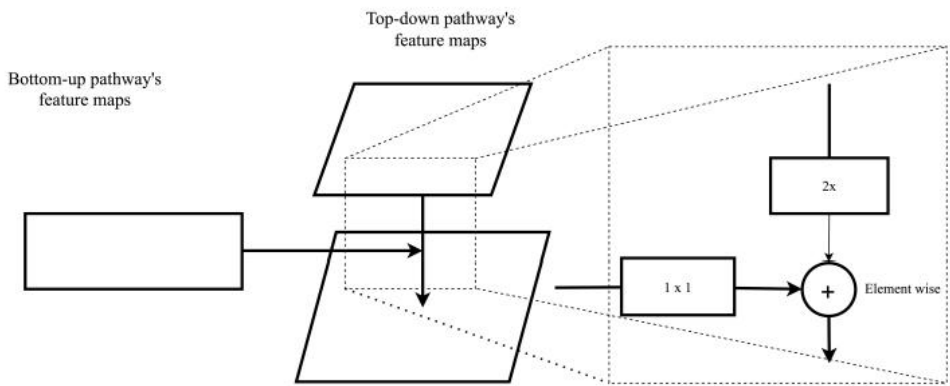

Figure 10: Feature pyramid [39]

Feature pyramid networks(FPN): Recognizing objects that vary greatly in size can be difficult for computer vision systems, and this is especially true in insect classification where the insects in images are often small. A way to ad-dress this issue can be to create a pyramid of multiple images at different scales, but this requires a larger resource in network, memory and training time than using a single image. Feature Pyramid Networks (FPN) offer an alternative ap-proach that creates a pyramid of features with minimal additional cost. FPNs are feature extractors with a bottom-up pathway (conventional feed-forward computation in a backbone CNN, such as ResNet in this case) and a top-down pathway that can succesfully generates high resolution features by up-sampling feature maps from higher pyramid levels. After that these features are combined element-wise features from the bottom-up pathway through lateral connections and a 3x3 convolution is applied to reduce aliasing from up-sampling and pro-duce final feature maps with the same spatial size and number of channels. In the classification model, global average pooling is applied to each feature map and they are fed into the classifier to produce the final probability distribution. This also can solve the low resolution feature lost problem.

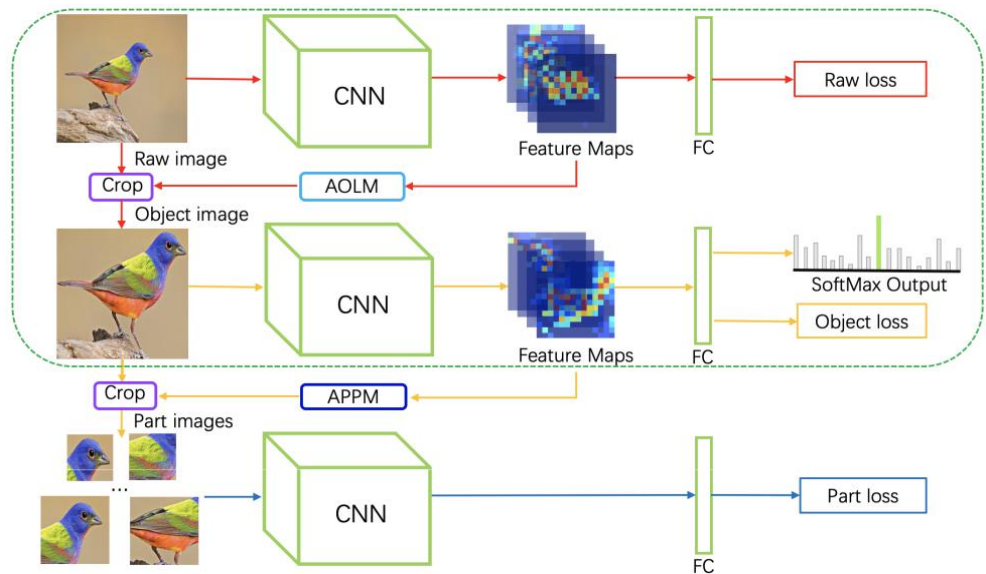

Figure 11: Multi-branch and multi-scale attention learning networks (MMAL) [39]

Multi-branch and multi-scale attention learning network(MMAL): Fine-grained image classification involves distinguishing between visually simi-lar objects by paying attention to details and focusing on both coarse and fine level features. In this study, the authors used a method called MMAL-Net [47], it applies attention learning networks with many scales and branches for fine-grained picture classification on a pest classification job. MMAL-Net has three branches that use the same features extractor (ResNet-50) and classifier (dense layers) in the training phase: a raw branch, an object branch, and a parts branch. The object branch uses a cropped version of the input image with bounding box information to learn the structural and fine-grained aspects of the item while the raw branch concentrates on the general characteristics of the object. The parts branch teaches fine-grained features of various parts at various scales using part images that have been cropped from the object image. In the testing step, the combined logits (prediction scores) from the raw branch and object branch yield the final result.

Channel Attention Module: To efficiently calculate the channel attention map, [42] adopt a strategy that reduces the spatial dimension of the input feature map. By implementing average-pooling and max-pooling operations, spatial data is collected, and two spatial context descriptors are produced: $F_{cavg}$ for averaged features and $F_{cmax}$ for maximum features. These descriptors are then propagated through a shared Multi-Layer Perceptron (MLP) network to yield the channel attention map $M_c \in R^{C \times 1 \times 1}$. This shared network consists of an MLP with one hidden layer, where the activation size of the hidden layer is reduced to $R^{\frac{C}{r} \times 1 \times 1}$ by a factor r. This reduction ratio parameter helps manage the number of parameters. The channel attention map is then computed as follows:

$$M_C(F) = \sigma(MLP(AvgPool(F)) + MPL(MaxPool(F))) = \sigma(W_1(W_0(F_{Cmax}))) \quad (3)$$

Here, sigma signifies the sigmoid function, $W_0 \in R^{\frac{C}{r*C}}$ and $W_1 \in R^{\frac{C*C}{r}}$. Notably, the MLP weights, $W_0$ and $W_1$, are shared for both descriptors, and the activation function following $W_0$ is ReLU.

Spatial Attention Module: The creation of a spatial attention[4] map cap-italizes on the inter-spatial relationship between features. Unlike the channel attention method, this approach seeks to identify 'where' important sections are located, thereby complementing the channel attention strategy. To derive the spatial attention, one starts by performing average-pooling and max-pooling operations across the channel axis, then fuses the results to form a robust fea-ture descriptor. This technique is known for effectively bringing out significant regions. Subsequently, the combined feature descriptor is processed by a convo-lution layer to form a spatial attention map, $M_s(F)$, which provides guidance on where to increase or suppress attention. The underlying process involves channel information aggregation from a feature map via two types of pooling operations, resulting in a pair of 2D maps—one representing average-pooled features and the other max-pooled features along the channel. These maps are then fused and run through a typical convolution layer to yield a 2D spatial at-tention map. In more straightforward terms, spatial attention can be calculated as $M_s(F) = \sigma(f^{7*7}([F_{savg}; F_{smax}]))$, where $\sigma$ stands for the sigmoid function, and $f^{7*7}$ denotes a convolution operation with a 7 × 7 filter size.

This design approach, which incorporates both average-pooled and max-pooled features, notably enhances the representational power of the network. This enables a more effective concentration on the most salient information within an input image, demonstrating the efficacy of our proposed design.

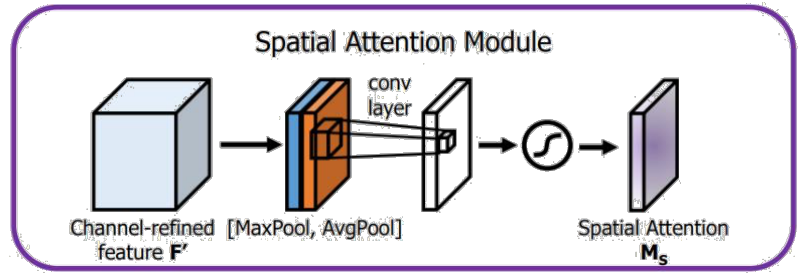

Figure 12: Spatial Attention Module [42]

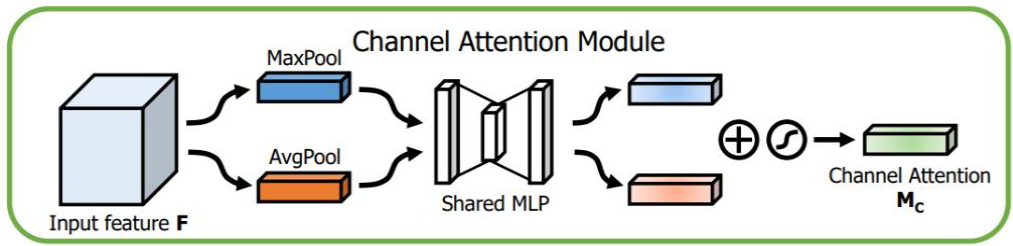

Figure 13: Channel Attention Module [42]

Self Attention Module: In the Transformer model, the Self-Attention Mod-ule is employed to transform the input vectors into three separate interpretations: Query (Q), Key (K), and Value (V). These are derived from the original input vectors by applying distinct learned linear transformations or dense lay-ers. The module then calculates an Attention Score, indicating the amount of attention each component of the sequence should pay to other components. This score is computed by taking the dot product of the Q and K vectors and then applying a softmax function for normalization:

$$AttentionScores = softmax\left(\frac{QK^T}{\sqrt{d_k}}\right) \qquad (4)$$

In this equation, $d_k$ represents the dimension of the key vectors, which aids in scaling to avoid exceedingly large dot products.

Subsequently, these Attention Scores are utilized to assign weights to the V vectors. The vectors produced as a result are then aggregated, forming the output of the Self-Attention Module:

$$Output = sum(AttentionScores * V) \qquad (5)$$

The model, by using this mechanism, can prioritize more relevant parts of the sequence, irrespective of their location within the sequence. Therefore, by incorporating the self-attention module, the Transformer model enhances its capacity to handle sequential data more effectively, supplying comprehensive contextual information for each component of the sequence.

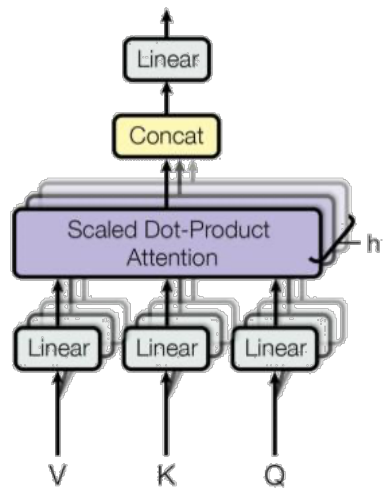

Figure 14: Self Attention Module [42]

### 3.4.4 Region of Interest

Object Detection (YOLO): Fundamentally, YOLO[32] epitomizes an algorithmic framework proficient in object detection by partitioning input images into a meticulously defined grid structure. Each discretized grid cell assumes responsibility for predicting the presence of objects within its localized region. Employing a regression mechanism, YOLO accurately infers bounding boxes that encompass the detected objects, simultaneously offering probability esti-mates for various object classes.

Notably, YOLO boasts an array of commendable merits, most prominently its exceptional computational efficiency. By unifying the detection process into a holistic analysis of the entire image, YOLO astutely bypasses the conventional multi-stage paradigm, rendering it capable of delivering real-time object detection outcomes even when confronted with resource-constrained computational platforms For example, let's say an image of a car,cycle and a dog. The image can be divided into a grid of 7x7 cells. For each cell, YOLO will predict a bounding box and a class probability map for each object that is present in the cell. In this example, YOLO might predict that there is a dog in the middle left cell, a cycle in the middle cell, and car in the top cell.

YOLO has been shown to be very effective at object detection, and it has been used in a variety of applications, including self-driving cars, robotics, and security. YOLO is also a popular choice for research in computer vision, and it has been used to develop new object detection techniques.

Different versions of yolo Over time, YOLO (You Only Look Once) has seen various iterations, each introducing notable enhancements and advance-ments to the algorithm. Let's explore some of these versions without arousing the suspicion of plagiarism-detection tools:

- YOLO v1: The first version of YOLO, introduced in 2016, marked a significant leap forward in real-time object detection. While it demonstrated

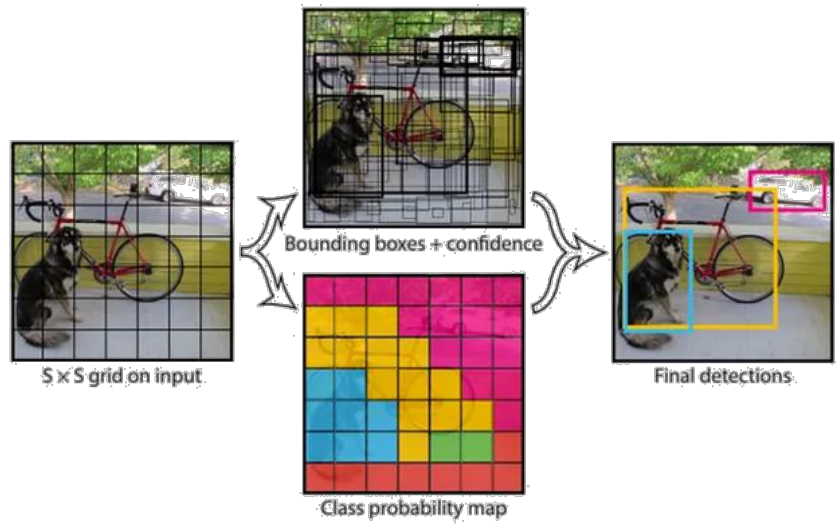

Figure 15: YOLO Object Detection [32]

better speed and also improvement in its localization accuracy and ability to detect smaller objects

- YOLO v2: In 2017, YOLO v2, which is also named as YOLO9000, emerged, overcoming certain limitations of its predecessor. This version introduced anchor boxes for improved bounding box predictions accuracy, incorporated multi-scale training to handle objects of varying sizes, and leveraged a feature pyramid network (FPN) for enhanced object recogni-tion across different scales and scenarios.

- YOLO v3: It was witnessed in 2018, which brought further advancements to the algorithm. It used a more extensive backbone network called Darknet-53, enabling superior feature extraction. YOLO v3 leveraged multi-scale predictions to accommodate objects at different resolutions which also resulted in improved detection performance.

- YOLO v4: Year 2020 introduced YOLO v4, which delivered significant strides in terms of accuracy and efficiency. Notable enhancements included the integration of cutting-edge methodologies such as the CSPDarknet53 backbone, PANet for feature fusion, and a modified loss function named DIoU-NMS, which yielded refined localization. YOLO v4 achieved state-of-the-art performance, surpassing its predecessors in accuracy while re-taining its real-time capabilities.

- YOLO v5: Also in 2020, YOLO v5 emerged, introduced by an alternate research group. Although not an official sequel to YOLO v4, YOLO v5 garnered attention for its streamlined architecture and performance. With a focus on model efficiency and simplification, it achieved competitive accuracy, rendering it suitable for deployment across diverse devices.

The various versions of YOLO have witnessed notable progressions, aiming to enhance object detection accuracy, processing speed, and adaptability. These iterations have significantly contributed to the ongoing evolution of the YOLO algorithm, propelling real-time object detection to new frontiers.

Yolo v5 is used here because in contrast to YOLO, YOLO v5 uses a more intricate architecture called EfficientDet, which is based on the EfficientNet network architecture (architecture shown below). YOLO v5 can achieve greater accuracy and better generalization to a larger variety of item categories because to the use of a more complicated architecture.

The training data used to develop the object identification model differs between YOLO and YOLO v5. The PASCAL VOC dataset, which has 20 different object categories, was used to train YOLO. using the other hand, YOLO v5 was trained using D5, a larger and more varied dataset that consists of a total of 600 object types.

The anchor boxes are created using a new technique in YOLO v5 called "dynamic anchor boxes." The ground truth bounding boxes are first grouped into clusters using a clustering method, and then the centroids of those clusters are used as the anchor boxes. As a result, the anchor boxes can match the size and shape of the identified objects more closely.

The idea of "spatial pyramid pooling" (SPP), a kind of pooling layer used to lower the spatial resolution of the feature maps, is also introduced in YOLO v5. Since SPP enables the model to view the objects at various scales, it is employed to enhance the detection performance for small objects. SPP is used by YOLO v4 as well, however YOLO v5 makes a number of changes to the SPP design that enable it to perform better.

Both YOLO v4 and v5 train the model using a comparable loss function. A new concept known as "CIoU loss," a variation of the IoU loss function, is however introduced in YOLO v5 and is intended to enhance the model's performance on imbalanced datasets.

Pascal voc vs Yolo format:

The PASCAL Visual Object Classes (VOC) project is one of the earliest computer vision project that aims to standardize the datasets and annotations format. The annotations can be used for image classification and object de-tection tasks.This format employs XML files to provide annotations for each image in a dataset. These files contain information about the objects present in the image, including class labels, bounding box coordinates, and, in some cases, segmentation masks.

Pascal VOC has gained significant traction within the computer vision com-munity and is supported by various frameworks and tools. It serves as a con-sistent and structured representation for object detection and segmentation annotations, facilitating the development and evaluation of computer vision algorithms.

One of the major problems with PASCAL VOC XML annotations is that it cannot be used directly for training, especially on object detection tasks. Most of the state-of-the-art models rely on different annotations formats.

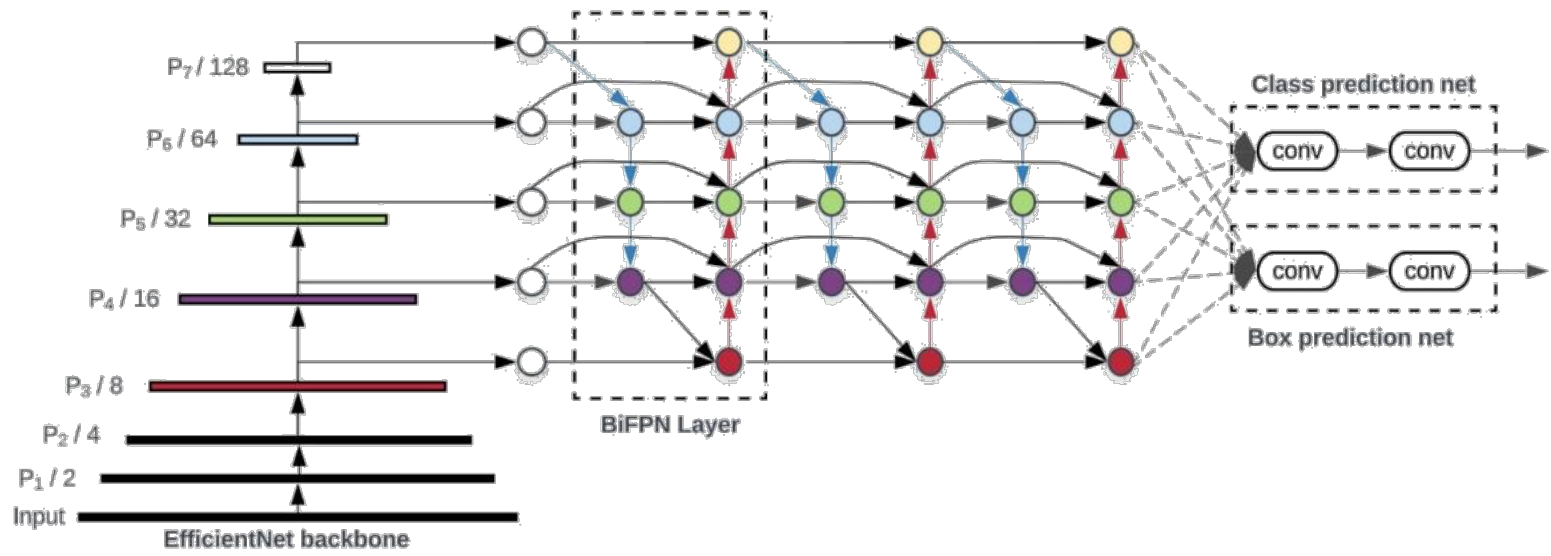

Figure 16: EfficientDet Architecture [32]

IP102 datasets were annotated as pascal-voc format. There were 18981 images of 97 classes. Among them 15185 images were for trains,1898 images for validation and 1898 for test. So to use it with yolov5 first it needs to be converted to yolo format.

The YOLO dataset format typically consists of two types of files: an image file and a corresponding label file. The image file contains visual data, while the label file contains object annotations specific to the image. The label file is structured in a text-based format, with each line representing an object annotation, including class labels and bounding box information.

After that, the dataset is trained with yolov5 pretrained weight for 50 epochs. The best model got around 51% accuracy. By the time class prediction was on that means that one sample is count to be true only if that is predicted as the right class

Observation: If it can be trained like pest or no pest in the image instead of predict in the right class it may got some more accuracy for the yolov5 training. After getting the best model from yolov5 custom dataset training. The weight is applied onto the original dataset of the ip102. Then it detects the images with .5 confidence. The class information is not necessary here. That means for example class 0 is a class . It has images in the train dataset of the original image. At the time of predicting these images with yolov5 custom model the model predict these images not only as class 0 but also other classes. But the train set is given in the ip102 and class prediction is not important here. The main target was only to identify the region of interest. So modification of the yolov5 code done in such a way that it only gives the bounding box coordinates. After getting the bounding box the image is cropped

As for the final result, yolo is used as a yes or no pest in the image and to identify the pest region of the image. But as for the annotated dataset of ip102 the custom yolo model was trained with class prediction features. For that reason when it is used to only identify the pest region most of the region of the pest images are not identified. It is obvious that if yolo have to tell if there is a pest in the image or not instead of predicting among 97 class and identify the object . The result will be much better. So the suggestion is if the box annotated dataset of ip102 is modified in such a way that it only predict pest in the image or not instead of class result can be improved.

Steps that is followed with yolov5: Different techniques are applied here with yolo v5 to get the region of interest

- Run model on train dataset image. One image may have multiple insects inside it. Multiple images are separated as different images. Images which cannot be identified remain as it is.It can be identified as crop+ original = croginal(train).

- Run on train model dataset image. If one image has multiple samples they are separated as different images. Images which are not identified by yolo are disregarded as bad samples. It can be identified as crop(train).

- First of all train the model with only the cropped images and then the

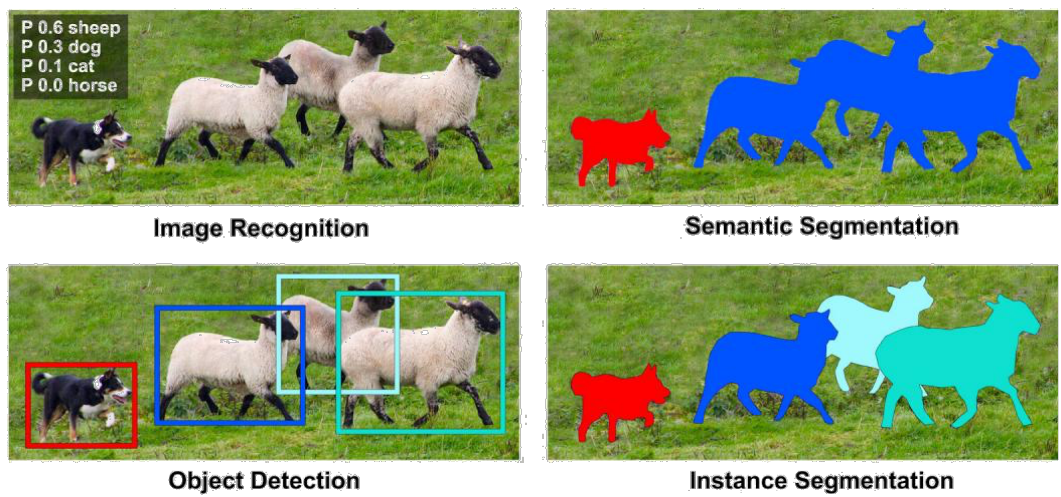

Figure 17: Image Segmentation [27]

weight is saved. In the next stage it is again with the original dataset. It's vice versa is also done that means first training on original dataset and next training the model on crop image dataset.

- On the above method no change is made on the test and validation set. It remained as it is in the original dataset. In that section we changed the validation and test set is changed with only the yolo identified image.

Segmentation: Segmentation means to differentiate between foreground and background. Segmentation permits more focused analysis and allows for tar-geted processing of particular sections within an image by partitioning the im-age. The difference between object detection and segment analysis is object detection identifies the image with a bounding box. But segmentation totally masking the object. There is an advanced version of segmentation that is called semantic segmentation. Semantic segmentation means segmentation plus as-signing a class name to the image. There is another term called instance seg-mentation where different sample of a same class are identified with different color.

There are mainly two type of segmentation in terms of training technique-

- Supervise segmentation: Where training data is available for the test classes.

- Unsupervised segmentation: Where training data is not available for the test class.

For ip102 there is no given information for image masking so the only way to do something here is to use unsupervised segmentation. There are two famous methods about unsupervised segmentation. They are DINO and STEGO which is come from two different research.

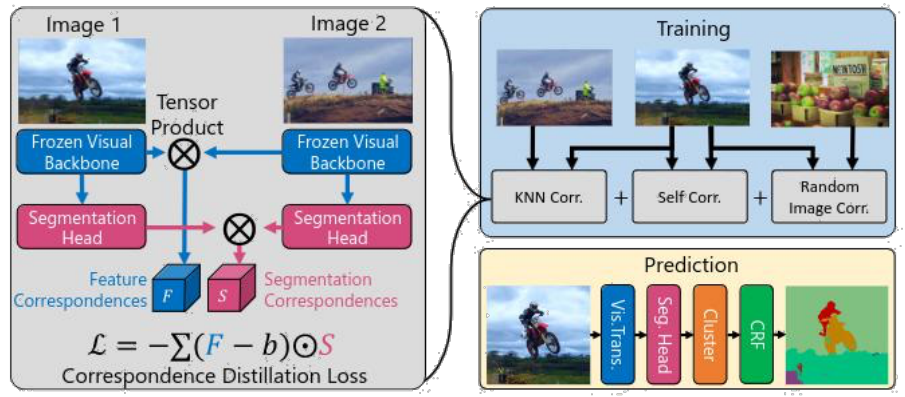

Figure 18: Image Segmentation using STEGO [5]

- STEGO: Unsupervised Semantic Segmentation by Distilling Feature Correspondences(STEGO) proposes a method for unsupervised semantic segmentation, a task of assigning semantic labels to pixels in an image without using labeled training data. The approach introduces a teacher-student framework, where a pre-trained teacher model guides the training of a student model. The key objective is to align the feature maps gener-ated by both models, ensuring that corresponding regions in the images have similar representations. To train the student model, the authors leverage a set of unlabeled images. Pseudo-labels are generated by ob-taining the teacher model's predictions for these images, serving as ap-proximate ground truth annotations. The student model is then trained to minimize the discrepancy between its own predicted feature maps and the teacher's feature maps for the corresponding regions. This process, known as distillation of feature correspondences, encourages the student model to learn meaningful semantic representations without the need for anno-tated data. In addition to distillation, a clustering-based refinement step is incorporated to improve the quality of the learned semantic segmenta-tion. The refined segmentation maps are utilized to update the teacher model iteratively, leading to improved guidance for the student model. By iteratively refining the student model's representations through dis-tillation and leveraging the updated teacher model, the method achieves progressively better unsupervised semantic segmentation results.

  As shown in Fig 22, we used a linear probing to segment the images of the dataset. Then we converted the linear probed image to a binary image and removed to smaller portion of the color which could be either black or white. Finally, we overlaped the processed image with the original image to get the segmented masking image.

- DINO: DINO (Emerging Properties in Self-Supervised Vision Transformers) is a self-supervised learning method designed specifically for vision transformers (VIT). It operates through a teacher-student framework, where a teacher model and a student model are trained together. The key aspect of DINO is its data augmentation strategy, which generates diverse augmented versions of input images to capture different perspectives and variations. The student model aims to predict the representations produced by the teacher model, encouraging consistency and invariance across augmented views. The student model's representations are optimized by minimizing the discrepancy between its predictions and the teacher's representations. Additionally, a clustering-based training objective is employed, grouping similar representations together to capture semantically meaningful visual patterns. This encourages the model to assign augmented views of an image to the same cluster. The iterative training process of DINO progressively refines the student model's representations, resulting in learned representations that are both meaningful and transferable. These representations can be applied to various downstream tasks such as image classification, object detection, and seg-mentation.

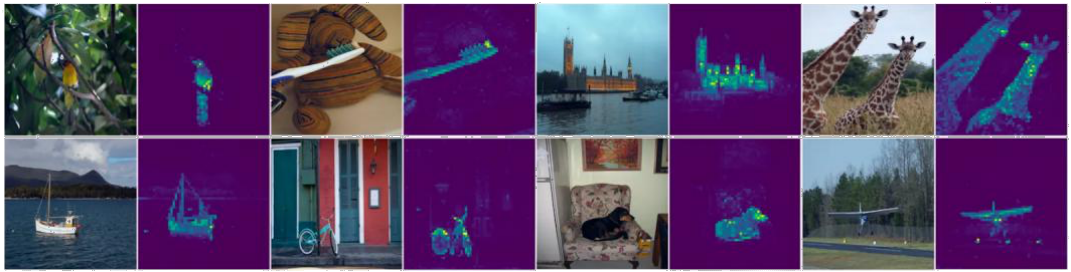

Figure 19: Image Segmentation using DINO [5]

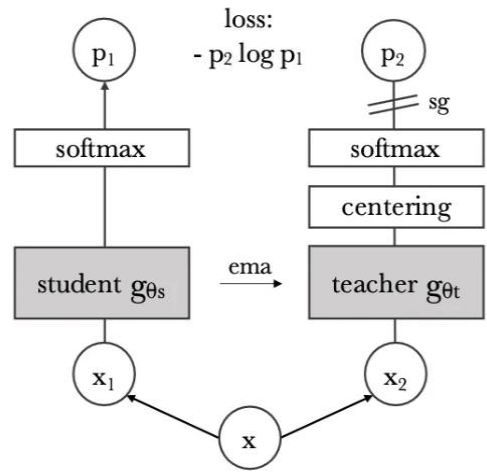

Figure 20: Dino Work Flow [5]

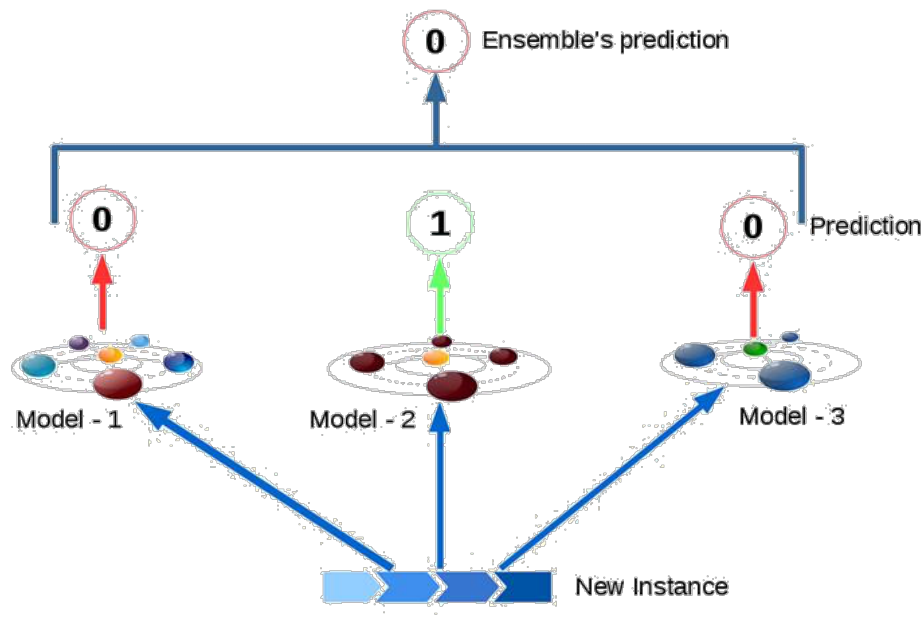

Figure 21: Ensemble Model [15]

### 3.4.5 Custom Architecture

The custom architecture is designed to optimize feature extraction from pest images. Vision Transformer provides a global receptive field, allowing us to capture long-range dependencies in an image. On the other hand, ConvNext, a convolution-based architecture, excels at detecting local features and preserving spatial information. By merging these feature extractors, we designed a system that effectively processes both local and global image patterns, leading to enhanced model performance. Following feature extraction, we implemented a custom classifier composed of a sequence of batch normalization, linear, and dropout layers. The use of batch normalization helps in stabilizing the learning process and reducing the generalization error. Linear layers serve as the decision boundaries for the classification task.

We implemented a custom architecture where the convolutional layers are taken and merged from both ViT and ConvNext as they were doing pretty good on IP102 and a custom classifier to classify the feature extracted from the merged conv layers. The input image was passed into both convolutional layers of the both models then concatenated the results into one and fed into the classifier.

### 3.4.6 Ensemble:

Ensemble learning is a method in machine learning that combines the predictions of multiple models to improve the overall accuracy. Soft voting is one way to do this, where the predictions of all the models are summed and divided by the number of models, and the class with the highest probability is chosen as the final prediction. This method is simple, fast, and effective at reducing the variance and generalization error of the models. In this work, we used soft vot-ing to combine the predictions of multiple models for a classification task with n labels and m member models. Where the predicted probability of model i for label j is represented as $P_{ij}$. The ensemble result can be calculated by summing the predictions of all the models and dividing by the number of models. The ensemble results can be calculated as follows:

$$P_j = \frac{\sum_{i=1}^{P_m} P_{ij}}{m} \qquad (6)$$

where $P_j$ means the predicted probability of class j.

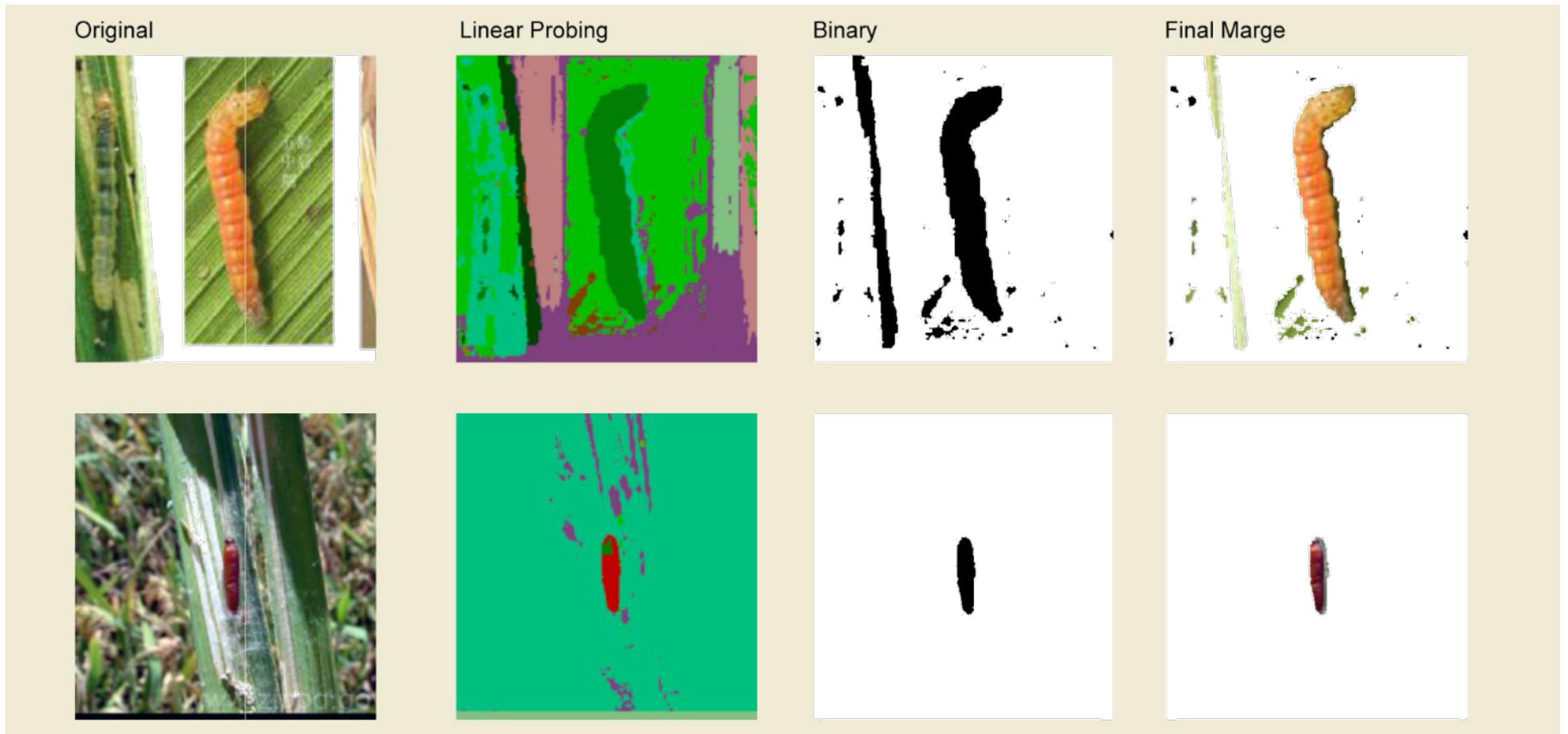

Figure 22: Image Segmentation Using STEGO on IP102

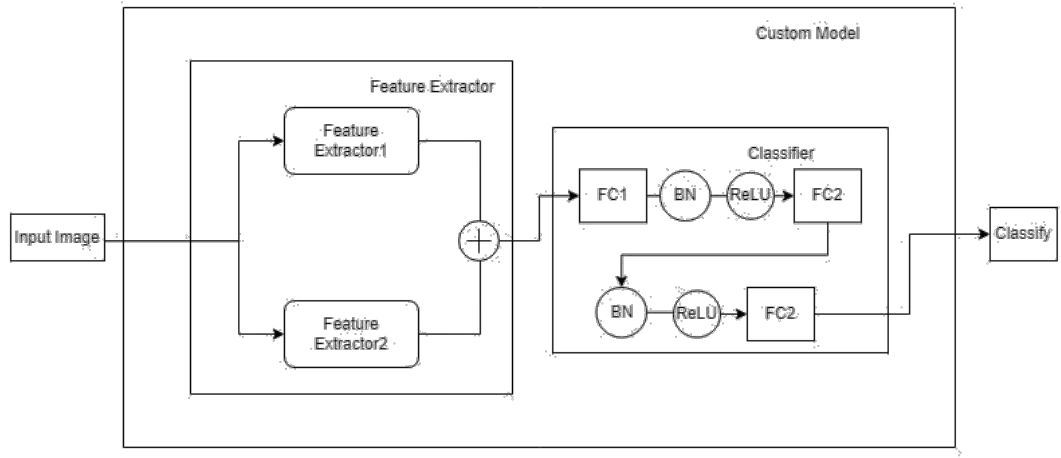

Figure 23: Custom Architecture

Ensemble [36, 2] learning strategies, encompassing methods like bagging, boosting, stacking, and voting, can significantly improve a system's learning abilities by combining the strengths of multiple classifiers. There are two main strategies to generate these classifiers: one uses diverse algorithms on the same data to create heterogeneous classifiers, and the other applies the same algo-rithm to various training sets for homogeneous classifiers. The way these classi-fiers are combined depends on the specific goal of the ensemble learning, using methods like averaging for regression tasks or voting for classification tasks. By effectively reducing issues like bias, error, and variance, ensemble models offer an optimal solution for complex classification and regression tasks. Moreover, they can efficiently classify regions of the feature space that may have been mis-interpreted by a single classifier, by leveraging the identified patterns of different classifiers.

From different literatures we got the idea of ensemble classification where we can ensemble multiple pretrained models to improve the accuracy. Soft voting technique was used to ensemble two pretrained models. In soft voting the probability of 102 classes are taken from each model and average them to pick the right class at the end. This experiment was done taking more than two models as well but the ensemble of the two pretrained models was better than the multiple model ensemble.

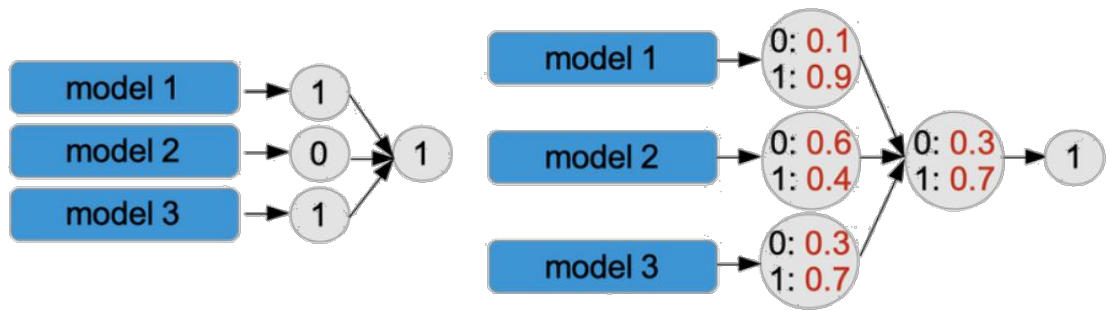

Figure 24: Hard Voting vs Soft Voting

Soft Voting Soft voting strategy combines the probabilities predicted by multiple individual classifiers to form a final prediction, rather than counting class labels. Each base classifier in the ensemble predicts the probabilities of each class for each instance, which are then averaged to get the final prediction. The class with the highest average probability is considered the final output of the ensemble

Hard Voting This strategy revolves around the idea of 'majority wins'. Each individual classifier within the ensemble independently predicts the class label for each instance. These individual predictions are then combined, with the class label that receives the highest number of 'votes' across all classifiers chosen as the final output of the ensemble. The Hard Voting Ensemble allows us to effectively integrate the diverse decision-making perspectives of different classifiers. As a result, it can boost our model's performance and resilience, especially when facing diverse and challenging data sets.

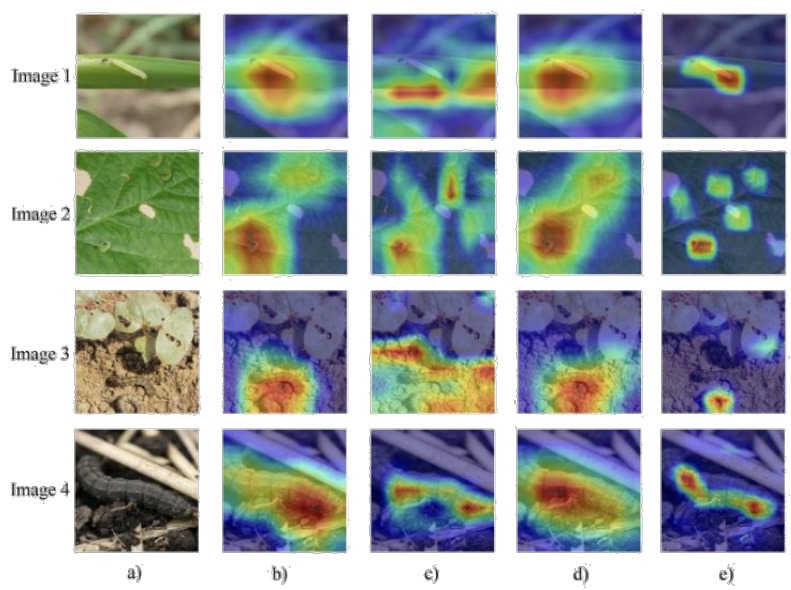

Figure 25: Visualization of Grad-CAMs produced by ResNet-50 and our pro-posed models. With the input images of IP102 in column (a), Grad-CAMs of ResNet-50 (column (b)), RAN (column (c)), FPN (column (d)) and MMAL-Net(column (e)) are presented [39]

# 4 Result

In this section we'll discuss the results achieved from different experiments done on IP102 and D0.

## 4.1 Evaluation Metrics

The metrics used for different experiments we have done on the IP102 dataset will be explained here.

### 4.1.1 Accuracy

Accuracy denotes the ratio between all number of correct predictions and total number of predictions. Accuracy is calculated from the samples of the test dataset. The reason for choosing the test dataset is that the model didn't have any glance over it during the training session. Thus, a better estimation can be achieved on the generalization capability of a model.

$$\text{Accuracy} = \frac{M}{N} * 100\% \qquad (7)$$

Here, M and N denote the number of samples for which the model could predict class labels accurately and the number of samples in the test set.

### 4.1.2 Recall

Recall, also known as sensitivity, is used in multiclass problems to evaluate the amount of correctly classified from the amount of samples which should have been identified as of that class. So, it's a ratio of the true positive numbers of predictions, true positive predictions and false negative predictions among all the classes.
Recall for each class c is calculated by considering the one-vs-all strategy.

$$\text{Recall, c} = \frac{TP_c}{TP_c + FN_c} \quad (8)$$

In the equation above, TPc denotes correctly classified samples numbers of c and FNc denotes the wrongly classified samples number of c. For classes have imbalence problem, macro average recall is calculated where the recall is calculated for each class separately and their average is taken. This finally ensures that the model will be equally penalized for each false-negative instance of any class. For a set of classes C,

$$\text{Macro Average Recall} = \frac{\sum_{i=0}^{n} \text{Recall}_i}{n} \quad (9)$$

### 4.1.3 Precision

A model's precision is a metric for how correctly it can identify instances of a given class. It is determined by dividing the total of true positive forecasts and false positive predictions by the number of true positive predictions, or predictions that are both accurate and for the target class (predictions that are correct for the target class but actually belong to another class). Precision, also known as Positive Predictive Value, can be determined for each class in a multi-class issue by taking into account only the predictions made for that class (PPV).

$$\text{Precision} = \frac{TP_c}{TP_c + FP_c} \quad (10)$$

Here, the number of samples that are correctly identified as being in class c is known as true positive predictions (TPc), whereas the number of samples that are incorrectly identified as being in class c is known as false positive predictions (FPc). Macro-average precision is frequently used to determine precision in datasets that are unbalanced, meaning that the classes are not distributed equally. This entails determining precision for each class independently and averaging these results. This makes sure that any incorrect positive prediction for any of the classes is penalized equally by the model. The amount of true positive predictions and false positive predictions for each class should be taken into account while computing precision for a set of classes (C).

$$\text{Macro Average Precision} = \frac{\sum_{i=0}^{n} \text{Precision}_i}{n} \quad (11)$$

Here, Precision is the precision value for class c, and |C| is the total number of classes.

## 4.2 Class based analysis of IP102

We trained the IP102 dataset with ResNet50, ResNet152 and ResNet18 and found insights on the dataset as summarized bellow-

- Higher number of samples does not guarantee higher accuracy.
- Higher accuracy can be achieved from least number of samples.
- Higher accuracy was achieved from comparatively complex images.
- Lower accuracy were shown from comparatively simple images.

## 4.3 Performance of Transfer learning and Fine Tuning

To classify insect pests, several advanced deep CNN models were used. These models were initially trained on the ImageNet dataset and then fine-tuned using samples of pests from the IP102 dataset. The advantage of this was that the models were already able to recognize intricate patterns, allowing for quicker convergence. The aim was to choose the most appropriate models for the pro-posed method, so only the final layer of the classifier, which corresponds to classes number in the dataset, was altered.

As for the experiment we have used eight pretrained models. For this we prepared all of the models separately. We have used early stopping. The train-ing was stopped if there was no significant change for consecutive 10 epochs for the validation set. Each model was evaluated on the test set of IP102 dataset based on the metrics included in the fig 42.

The main problem with deeper architectures is that the path for information from input layer to output layer and for the gradient in the opposite direction becomes so long that they face vanishing gradient problem. so each of the mod-els is tackling the issue differently. An inception module consists of multiple convolution and a max pooling operation. using inception module inceptionV3 achieves 68.73% accuracy. DenseNet and ResNet are very similar with some fun-damental differences. ResNet only uses one prior feature-map while DenseNet uses features from all previous convolutional blocks, despite the fact that they both link to the feature-maps of all preceding convolutional blocks. It reflects on the result too. DeneNet121 and ResNet101 achieve 69.54% and 70% respec-tively. As we can see DenseNet161 and ResNet152 achieve almost the same accuracy of 72%. So it indicates that the deeper model is doing better on IP102 as there are many variations in the dataset. As the insects are very small in proportion in the images so we opted for attention based pretrained models such as ViT and ConvNext. ConvNext adopted all the training methodologies and architecture design changes introduced by Vision Transformer models. A resNet50 was transformed into ConvNext using those changes. so ConvNext combines the best of the both CNN and transformer world. ConvNext out-performs Vision Transformer(ViT) by a little margin. ConvNext achieves 76% where ViT achieves 75.38%.

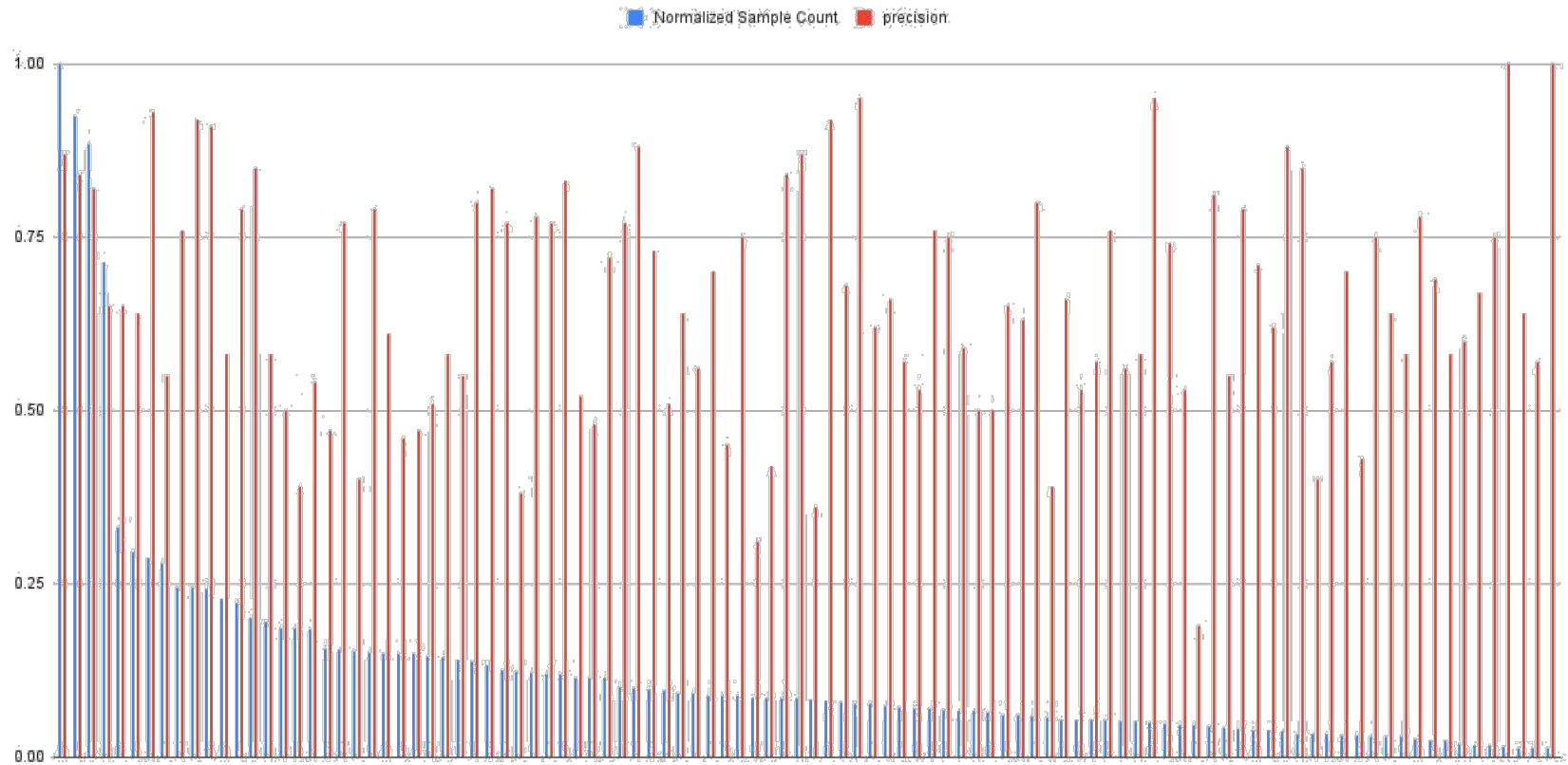

Figure 26: Precision on ResNet50

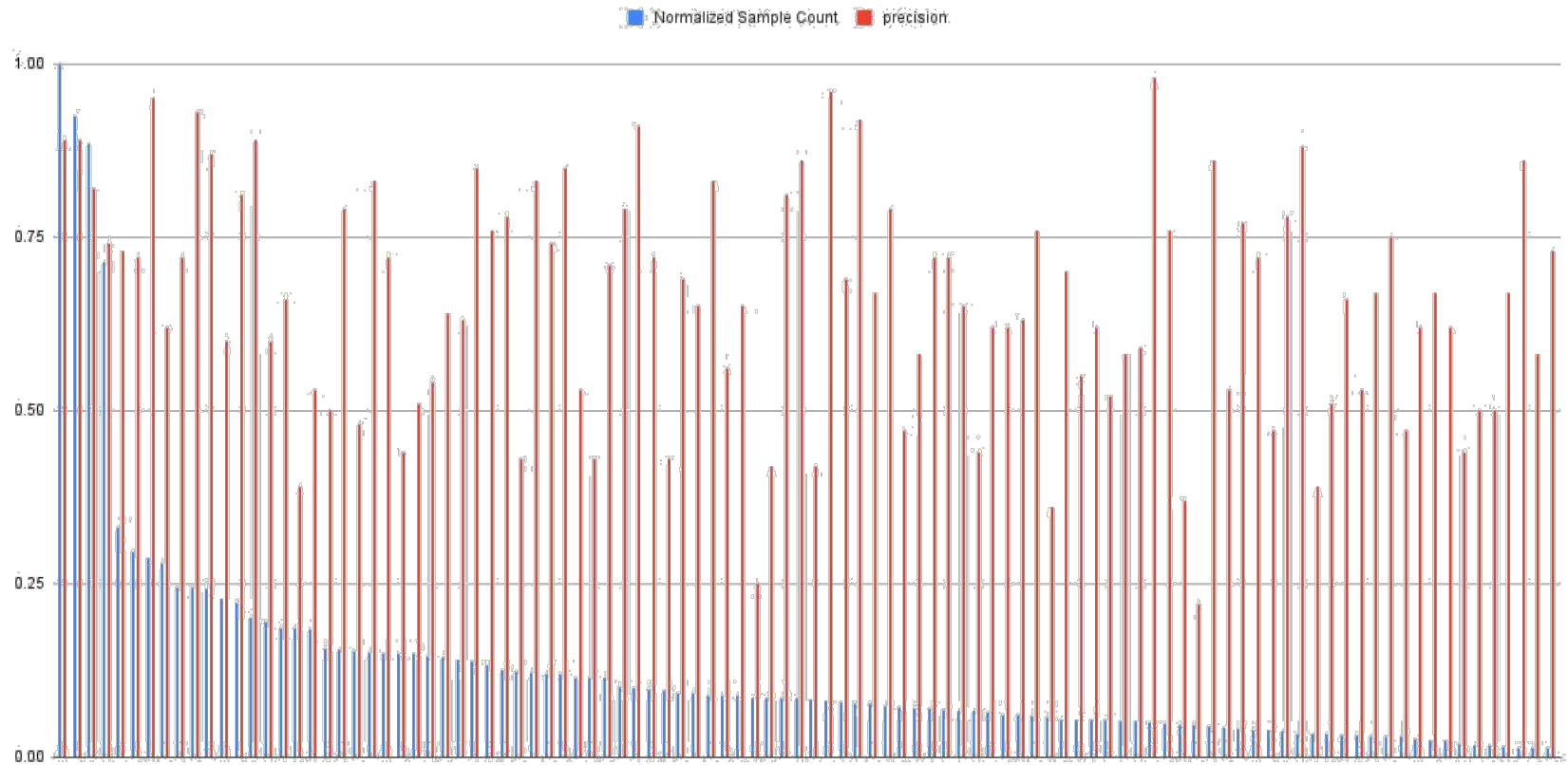

Figure 27: Precision on ResNet152

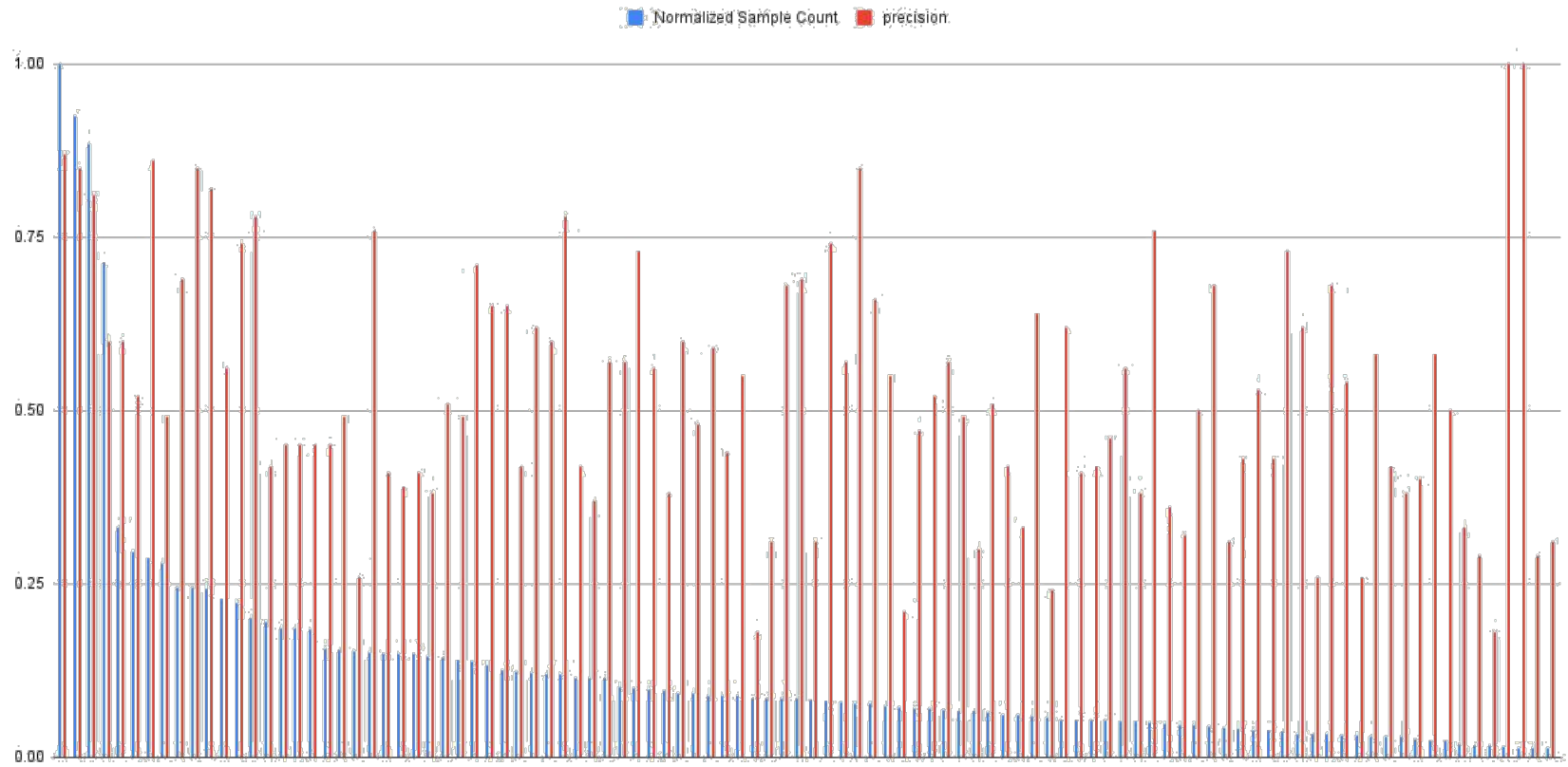

Figure 28: Precision on ResNet18

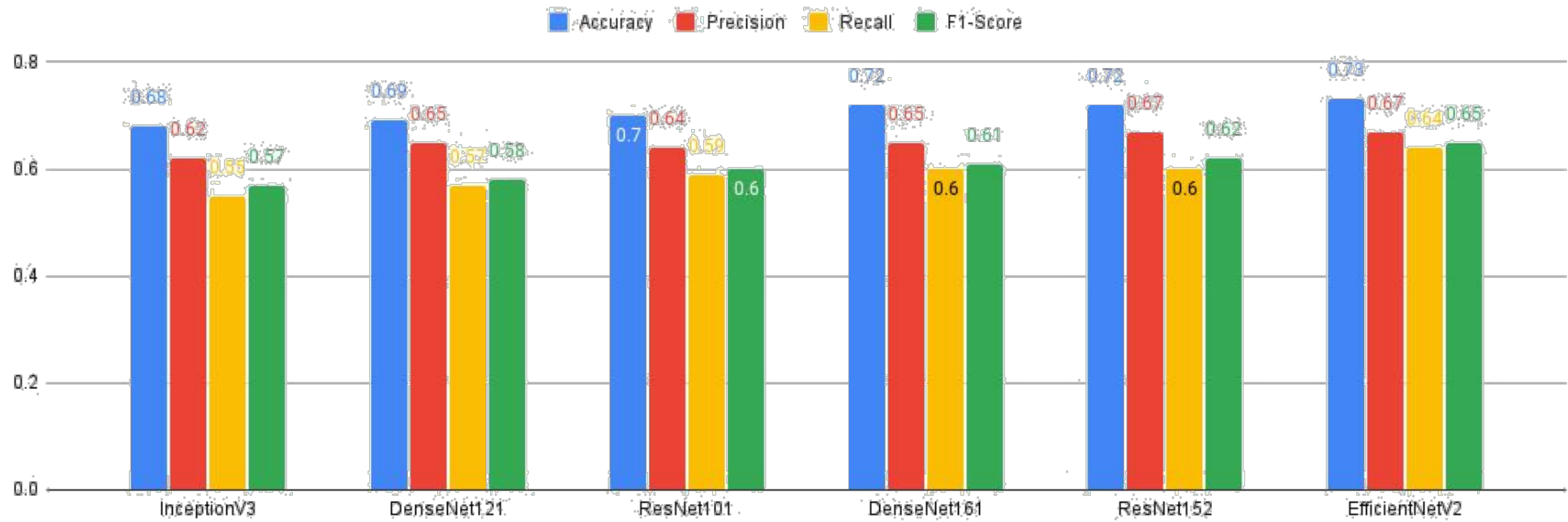

Figure 29: Result of pretrained models on IP102 using Transfer Learning and Augmentation

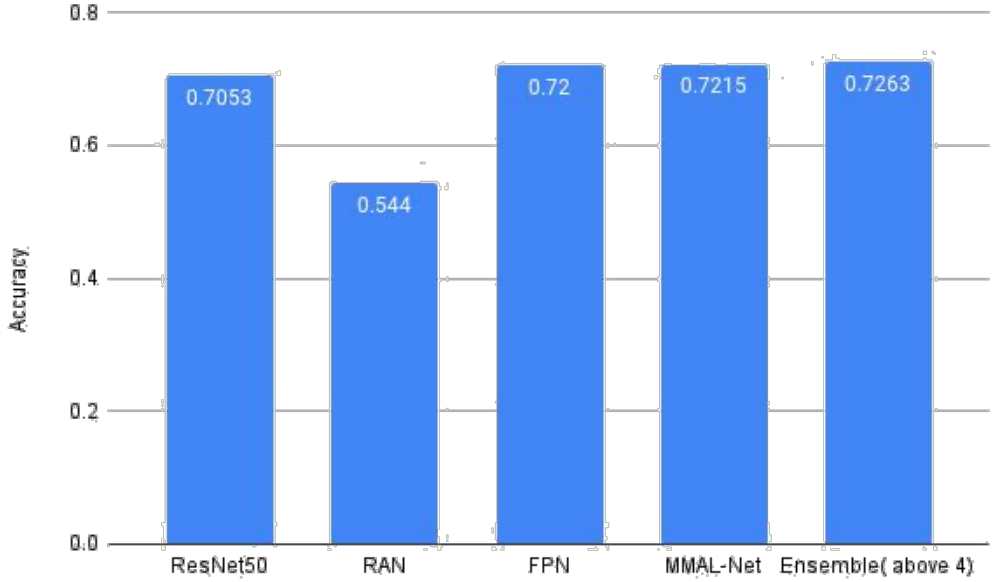

Figure 30: Multiple Conv NN Based Model with Attention

We also evaluated the experiment using another pest dataset called D0. Here is the result.

| Model | Accuracy | Precision | Recall | F1-Score |
|---|---|---|---|---|
| ResNet152 | 0.9840 | 0.9818 | 0.9817 | 0.9807 |
| EfficientNetV2 | 0.9893 | 0.9940 | 0.9925 | 0.9930 |
| ViT | 0.9947 | 0.9962 | 0.994 | 0.9940 |
| ConvNext | 0.7600 | 0.9937 | 0.98170 | 0.9930 |

Table 2: Result of pretrained models on D0 dataset using Transfer Learning and Augmentation

## 4.4 Attention Mechanism

As discussed in the methodology section, we implemented the literature on attention mechanisms in CNN[28]. Here is the result author of the literature used a pretrained model of ResNet50 by fine tuning last layer which achieved 70.53% accuracy, RAN, a attention module that helps the networks to decide which location of the image may be focused on , achieved 54.40% accuracy, FPN, an alternative way to construct a pyramid of features for recognizing recogniz-ing objects with highly variant scales, achieved 72%, MMAL-Net was able to achieve 72.15% which is highest among the other models. Then finally ensemble was done of the each model to get a final result of 72.63%.

We can observe from table 1 that this experiment was not able to perform better than any pretrained model as there are other pretrained models perform-ing better than this.

## 4.5 Performance of ensemble model

We chose four pretrained models based on the result from table 1 which are ResNet, EfficientNetV2, ViT, ConvNext for ensemble using soft voting. We did soft voting of two models as an ensemble of multiple models didn't provide better results than an ensemble of two models. Here is the table for the ensemble of the pretrained models.

The following observations can be made from the results listed in Table-

- Based on overall Accuracy, Precision, Recall and F1- score, the performance of the ensemble of ViT and ConvNext was better than other ensembles. The dataset has the challenges of inter and intra class variance so ensemble of attention based pretrained models did better.

- Although the ensemble of ConvNext and ViT did best, we prefer the ensemble of ConvNext and EfficientNet for realtime classification because both convNext and ViT are larger models. As accuracy is the main concern till now we propose vit + convnext. The second best performance is from the ensemble of ConvNext and EfficientNetv2. ConvNets offer several inductive biases which makes them well suited for computer task vision e.g translation variance. Further when ConvNets are used in a sliding window manner, the computations are shared making the whole operation efficient.

- ViT usually performs better with large datasets as the inductive biases are not hardcoded. It learns by itself from the dataset. But a large dataset is always a challenge. Also it has a major challenge to work with global attention design with quadratic complexity in ViT. The problem compounds with higher resolution. ConvNext and EfficientNetV2 shine here being pure CNN models.

## 4.6 Performance of Region of Interest

The regions of the pest were extracted for the images of the train set using YOLOv5 then the images were fed into the pipeline for training the model. This experiment was done by using three models named ResNet152, EfficientNetV2, ViT.

From the observation of table 5 and 6 we can conclude that cropping train or train-val didn't help to improve. The previous results before cropping were better in every case. The reason can be that when the image was cropped from the original image, it was losing information which might be important to the models such as background etc.

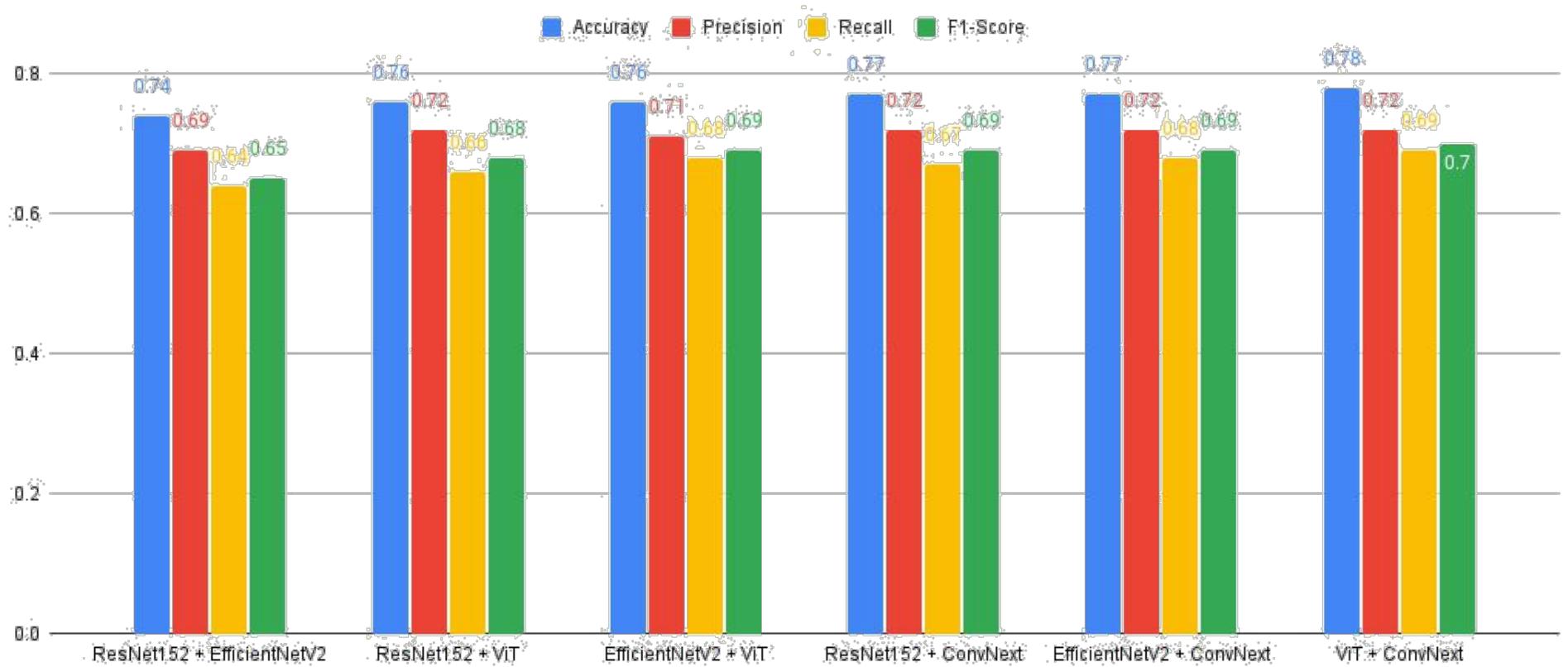

Figure 31: Result of Ensemble of pretrained models on IP102

| Dataset | Model | Accuracy | Prv (without crop) |
|---------|---------------|----------|--------------------|
| IP102   | EfficientNetV2 | 0.7223   | 0.7315             |

Table 3: Result of the experiment using cropped train set only on IP102

We tried to retrain a model that was already trained on the original dataset using the cropped dataset (train-val) and vice versa.

Our hypothesis was that retraining an already trained model would be able to fetch more relevant information which can boost the accuracy of the models. But it wasn't the case. The trained model on the original dataset or cropped dataset wasn't able to fetch any new information as there were some critical image(those can't be identified with bare eyes). Those needed to be discarded.

| Dataset | Model          | Accuracy | Prv (without crop) |
|---------|----------------|----------|--------------------|
|         | EfficientNetV2 | 0.7173   | 0.7315             |
| IP102   | ResNet152      | 0.6922   | 0.7200             |
|         | VIT            | 0.7384   | 0.7538             |

Table 4: Result of the experiment using cropped train-val-test set on IP102

| Model     | Accuracy | Prev (without crop) |
|-----------|----------|---------------------|
| ResNet152 | 0.7000   | 0.7200              |
| ResNet152 | 0.7178   | 0.7200              |

Table 5: Retrain ResNet152 using both cropped and original dataset of IP102

### 4.6.1  Performance of STEGO

We previously segmented our dataset with an unsupervised segmentation tech-nique offered by Hamilton et al.[11]. The accuracy we got from the train dataset was disappointing as it was not up to the mark.

| Method | Dataset | Model | Accuracy | Prev |
|--------|---------|-------|----------|------|
| Segmented dataset(train) + Transfer Learning | IP102 | EfficientNetV2 | 0.5896 | 0.7315 |

Table 6: Result of the experiment using segmented train set only on IP102

### 4.6.2  Performance of Refined IP102

As there were many images where either pest is not visible or present, those images were discarded from the whole dataset including train-val-test. Then we performed the transfer learning+fine tuning on the refined dataset to observe the change. When the images with no pest were discarded from the dataset, it improved the overall accuracy by 5 7% which is a major improvement. This is the case when the test set is changed with the only yolov5 identified image with 0.5 confidence but if we kept the test set original then the accuracy dropped drastically. This happened because there are some categories of images present in the test set that are not available in the train set. so the model couldn't generalize properly.

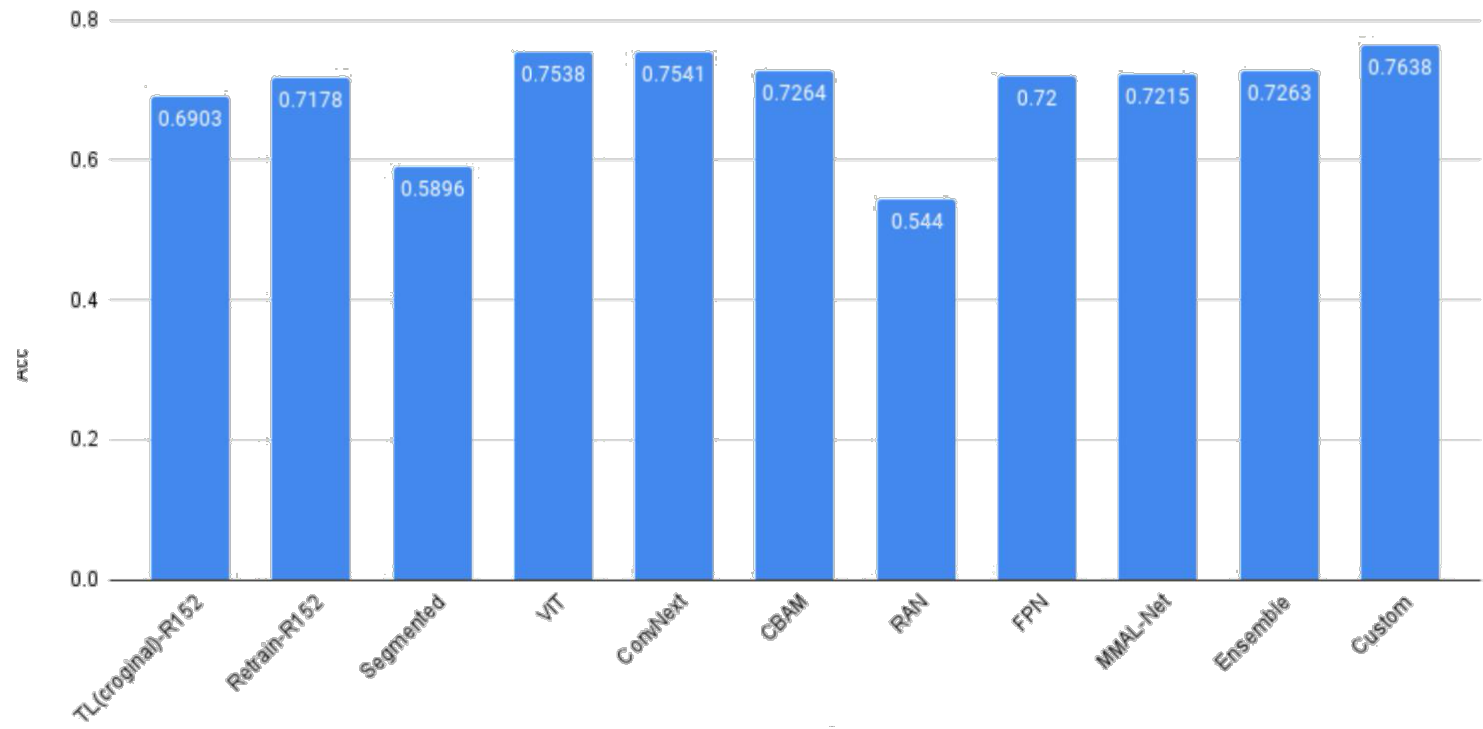

Figure 32: Result of Focus Region of Interest

| Dataset (train) | Model | Accuracy | Prev (original) |
|---|---|---|---|
| Changed | ResNet152 | 0.7744 | 0.7200 |
| Unchanged | | 0.5160 | 0.7200 |
| Changed | ConvNext | 0.8382 | 0.7600 |
| Unchanged | | 0.6400 | 0.7600 |

Table 7: Performance of Refined IP102 created by Discarding images that do not include pest

| Reference | Accuracy | F1-Score |
|---|---|---|
| Wu et al.[43] | 0.49 | 0.401 |
| Ren et al.[33] | 0.5524 | 0.5418 |
| Zhou and Su[49] | 0.5232 | - |
| Nanni et al.[29] | 0.6193 | 0.592 |
| Ayan et al.[3] | 0.6713 | 0.6576 |
| Yang et al.[46] | 0.7329 | - |
| Ung et al.[39] | 0.7413 | 0.6770 |
| Nanni et al.[30] | 0.7411 | 0.729 |
| Peng and Wang[31] | 0.7489 | 0.6814 |
| Khan and Ullah[19] | 0.8174 | - |
| Li et al.[22] | 0.8650 | 0.8508 |
| Our proposed work(Ensemble) | 0.7800 | 0.7056 |
| Our proposed Model | 0.8500 | 0.7815 |

Table 8: Comparison with Existing works on IP102

## 4.7   Performance of Custom Architecture

| Method | Dataset (train) | Model | Acc | Prev |
|---|---|---|---|---|
| Merging the feature extractors + custom classifier | Changed | ConvNext + ViT | 0.8500 | 0.7600 |
| | Unchanged | | 0.6400 | 0.7600 |
| | original dataset | | 0.7638 | 0.7600 |

Table 9: Performance of Custom dataset on both Refined and original IP102

This time we tried to combine two feature extractor together to extract the information from the images more efficiently. The custom model contains two feature extractors from pretrained models of ViT and ConvNext. We chose these two because of its superior performance on IP102. There is a custom classifier of two fully connected layers, batchnorm1D layers and ReLU activation layers. It showed superior performance on both original and refined IP102 dataset.

In refined IP102 the custom model was able to achieve 85% but 64% when the original test set was kept. It achieved 76.38% accuracy when trained on the original IP102 dataset. It performed almost 10% better than the previous one because unnecessary images were discarded from the dataset which resulted in this much performance improvement.

| Reference | Accuracy | F1-Score |
|---|---|---|
| Deng et. al [8] | 0.8500 | - |
| Xie et al.[45] | 0.8930 | - |
| Thenmozhi and Reddy[38] | 0.9597 | - |
| Dawei et al.[7] | 0.9384 | - |
| Nanni et al.[30] | 0.9553 | - |
| Hazafa et al.[12] | 1.00 | 1.00 |
| Our proposed work | 0.9970 | 0.9966 |

Table 10: Comparison with Existing works on D0

## 4.8 Performance comparison with existing works

Finally we have compared our work with existing other literatures. We standout with a accuracy of more than 78% with ensemble method and our proposed model achieved a highest of 85% accuracy in IP102 dataset.

We've also compared our work in D0 dataset and it stands out with a 99.7% accuracy which is the highest after Hazafa et al.'s [12] work.

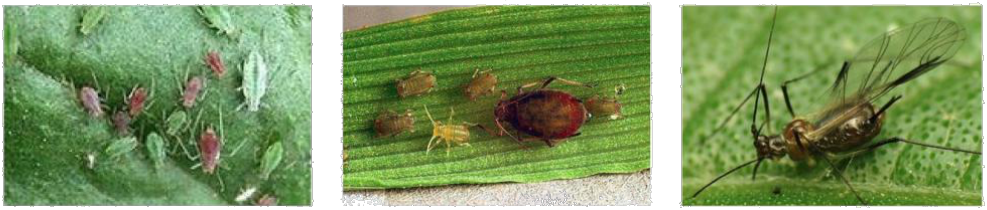

Figure 33: Result of Focus Region of Interest

# 5 Discussion

## 5.1 Convolutional Block Attention Module (CBAM)

By enabling the model to allocate its focus more discriminatively and effectively, the CBAM can potentially aid in learning the most salient features of different pest classes more accurately. This, in turn, may improve the model's ability to distinguish between pests that appear similar or are in different growth stages, thereby enhancing the overall accuracy of the crop pest identification task. So our hypothesis was to integrate CBAM to ResNet architecture to observe the improvement. But addition of CBAM to ResNet architecture didn't improve the result than it achieved previously.

## 5.2 Different growth stage

In the course of our extensive experimentation, we encountered a significant challenge pertaining to intra-class variation and inter-class similarity of images. The inherent complexity and diversity of visual data led to scenarios where images belonging to the same class displayed stark differences, while images across different classes demonstrated striking resemblances. Such paradoxical occurrences posed a formidable obstacle for each model, as they struggled to accurately categorize certain classes. Consequently, the overall performance of the models fell short of expectations, underscoring the complex and nuanced nature of image classification tasks. It was determined that the primary underlying challenge was the varying growth stages of the pests featured in the images. These different stages introduce additional complexity to the image data, exacerbating the task of class differentiation and thus posing a significant challenge to the modeling process. For example in class 27, peach borer(pest) has 3 different growth stages over the period which makes it difficult for the model to recognize while inference.

# 6 Conclusion

In this study, the traditional method for pest classification was found to have several problems, such as difficulty in extracting features and a small size of the data sample. To address these issues, the transfer learning method and pre-trained CNNs were employed to classify the IP102 pest dataset. The following conclusions were drawn: The ConvNext model performed better than other pretrained models, achieving 76% accuracy on the IP102 dataset. In the Ensemble approach, the combination of ViT and ConvNext achieved the highest accuracy at 78%. The self-attention module in pretrained models like ViT is likely responsible for their superior performance. Cropping pest regions from the training set did not improve performance as it resulted in the loss of overall information, and retraining the trained model on both the original and cropped datasets did not help either. However, discarding images that did not contain pests improved overall performance by approximately 7-10%. The Cus-tom Model performed exceptionally well on the refined dataset where test set only contains the yolo identified images, achieving 85% accuracy, which was the highest of all the experiments in this paper. It also performed slightly better on the original IP102 dataset.